\documentclass[journal]{IEEEtran}

\usepackage{forest}
\usepackage{tikz}
\usetikzlibrary{positioning, shapes.geometric, arrows.meta, shapes.arrows}
\usepackage{fancyhdr}
\usepackage{cite}
\usepackage{graphicx}
\usepackage{verbatim}
\usepackage{float}
\usepackage{caption}
\usepackage{amsmath,amsthm,amsfonts,amssymb,amscd}
\usepackage{balance}
\usepackage{totpages}
\usepackage{subfigure}
\usepackage{enumerate}
\usepackage{pifont}
\usepackage{authblk}

\usepackage{fancyhdr}
\usepackage{mathrsfs}
\usepackage{xcolor}
\usepackage{graphicx}
\usepackage{listings}
\usepackage{float}
\usepackage{textcomp}
\usepackage{xcolor}
\usepackage{mathrsfs}
\usepackage{hyperref}
\usepackage{ragged2e}
\usepackage{rotating}

\usepackage[utf8]{inputenc}
\usepackage{lipsum}
\usepackage{amsmath,bm}
\usepackage{indentfirst}
\usepackage{amsbsy}
\usepackage{cleveref}
\usepackage{mathrsfs}
\usepackage{enumitem}
\usepackage[ruled, vlined, linesnumbered]{algorithm2e}
\usepackage{algpseudocode}
\usepackage{multirow} 
\usepackage{booktabs}
\usepackage{pifont}
\usepackage{tabularx}
\makeatletter

\usepackage{soul}

\tikzset{
    base/.style={draw, thick, align=center, minimum height=2em},
    process/.style={base, minimum width=2.5cm, fill=orange!15},
    io/.style={base, fill=yellow!20},
    point/.style={coordinate},
    arrow/.style={->, thick, >=stealth},
    line/.style={thick}
}

\begin{document}

\title{Art and Science of Quantizing Large-Scale Models: A Comprehensive Overview}

\author[1]{Yanshu Wang\thanks{\texttt{yanshuwang@pku.edu.cn}}}
\author[1]{Tong Yang\thanks{\texttt{yangtong@pku.edu.cn}}}
\author[3]{Xiyan Liang}
\author[2]{Guoan Wang}
\author[4]{Hanning Lu}
\author[1]{Xu Zhe}
\author[5]{Yaoming Li}
\author[1]{Li Weitao}

\affil[1]{Peking University, Beijing, China}
\affil[2]{Beijing Institute of Technology, Beijing, China}
\affil[3]{Nankai University, Tianjin, China}
\affil[4]{University of Leeds, Leeds, United Kingdom}
\affil[5]{Harbin University of Commerce, Harbin, China}
\maketitle

\begin{abstract}
This paper provides a comprehensive overview of the principles, challenges, and methodologies associated with quantizing large-scale neural network models. As neural networks have evolved towards larger and more complex architectures to address increasingly sophisticated tasks, the computational and energy costs have escalated significantly. We explore the necessity and impact of model size growth, highlighting the performance benefits as well as the computational challenges and environmental considerations. The core focus is on model quantization as a fundamental approach to mitigate these challenges by reducing model size and improving efficiency without substantially compromising accuracy. We delve into various quantization techniques, including both post-training quantization (PTQ) and quantization-aware training (QAT), and analyze several state-of-the-art algorithms such as LLM-QAT, PEQA(L4Q), ZeroQuant, SmoothQuant, and others. Through comparative analysis, we examine how these methods address issues like outliers, importance weighting, and activation quantization, ultimately contributing to more sustainable and accessible deployment of large-scale models.
\end{abstract}


\IEEEpeerreviewmaketitle

\setlength{\subfigcapskip}{-0.2cm}
\setlength{\subfigbottomskip}{-0.15cm}
\setlength{\belowcaptionskip}{-0.3cm}

\section{Introduction}
\subsection{Background and Motivation}
\subsubsection{Machine Learning}

Machine Learning (ML) is a subfield of artificial intelligence that enables computers to learn from and make decisions based on data patterns without being explicitly programmed. The core of machine learning is developing algorithms that allow computers to receive input data and use statistical analysis to predict or classify outputs, thereby optimizing the performance of a specific task with minimal human intervention. Machine learning can be broadly categorized into three types:
\begin{itemize}
  \item \textbf{Supervised Learning:} The model is trained on a pre-defined set of data examples. The goal is to learn a general rule that maps inputs to outputs.
  \item \textbf{Unsupervised Learning:} The model looks for patterns and structures in data that is not labeled.
  \item \textbf{Reinforcement Learning:} The model learns through trial and error to perform a task to maximize the reward.
\end{itemize}

Machine Learning has many Applications: 

\textbf{Healthcare:}
In healthcare, machine learning applications include disease diagnosis, drug discovery, and medical imaging analysis. For instance, models trained to recognize pathological images can help doctors diagnose diseases like cancer more quickly.

\textbf{Financial Services:}Machine learning in finance is used for credit scoring, fraud detection, and algorithmic trading. It analyzes customers' transaction behaviors to identify fraudulent activities.

\textbf{Autonomous Driving and Robotics}\\
Autonomous vehicles use machine learning to process complex data from sensors to recognize the environment, make decisions, and increase driving safety and efficiency.

\textbf{E-commerce and Recommendation Systems}\\
E-commerce platforms utilize machine learning to analyze user behavior, optimize search results, and recommend products, significantly enhancing user experience and sales efficiency.

\textbf{Image Recognition}\\
Machine learning now identifies and classifies objects in images, widely used in social media, security surveillance, and industrial visual inspection systems.

\textbf{Natural Language Processing}\\
From speech recognition to text analysis, machine learning's natural language processing technologies allow machines to better understand and generate human language, used in chatbots, translation services, and sentiment analysis.

\subsubsection{Neural Networks the Evolution Towards Larger Models}
Neural networks, a pivotal concept in machine learning, are inspired by the biological neural networks that constitute animal brains. They are comprised of interconnected nodes or neurons, which collectively learn to perform complex tasks. Typically, these tasks include but are not limited to classification, regression, and pattern recognition, making neural networks versatile tools in both theoretical and applied machine learning.

\textbf{Structure of Neural Networks}\\
The basic structure of a neural network involves three types of layers:
\begin{itemize}
  \item \textbf{Input Layer:} This layer receives the raw input data analogous to sensory input in biological systems.
  \item \textbf{Hidden Layers:} One or more hidden layers compute functions applied to values from the previous layer. These layers form the core computational engine of the neural network.
  \item \textbf{Output Layer:} The final layer produces output for the network, which corresponds to the predictions for supervised learning tasks.
\end{itemize}

Each neuron in these layers applies a non-linear transformation to its input data and passes this output to the next layer. The strength and nature of the connections between neurons are adjusted through a process known as learning, typically implemented via backpropagation and gradient descent techniques.

\textbf{Learning Process in Neural Networks}\\
Learning in neural networks involves adjusting the weights of connections based on the error between the predicted output and the actual output. The most common learning algorithm used is backpropagation combined with an optimization technique such as gradient descent. This process involves:
\begin{enumerate}
  \item Propagating inputs forward through the network to generate the output.
  \item Calculating the error between predicted and actual outputs.
  \item Propagating the error backward through the network to update the weights, aiming to minimize the error by adjusting the weights.
\end{enumerate}

\textbf{Types of Neural Networks}\\
There are several types of neural networks, each designed for specific types of problems and datasets. These include:
\begin{itemize}
  \item \textbf{Convolutional Neural Networks (CNNs):} Highly effective for processing data that has a grid-like topology, such as images.
  \item \textbf{Recurrent Neural Networks (RNNs):} Designed to handle sequential data, such as time series or language.
  \item \textbf{Deep Belief Networks (DBNs):} A type of deep network that uses a stack of restricted Boltzmann machines layered on top of each other.
\end{itemize}

\textbf{Applications of Neural Networks}\\
Neural networks have been successfully applied in numerous domains including:
\begin{itemize}
  \item \textbf{Vision Systems:} From facial recognition to autonomous driving.
  \item \textbf{Speech Recognition:} Enabling voice-activated assistants and real-time translation services.
  \item \textbf{Natural Language Processing:} Driving the development of conversational AI and other language understanding applications.
\end{itemize}
Over the past few decades, there has been a significant shift in the architecture of neural networks, from relatively simple designs to highly complex and large models. This trend is driven by the continuous growth in computational power and the availability of vast amounts of data, which have enabled the training of larger models capable of performing a multitude of tasks with unprecedented accuracy.

\textbf{Evolution of Model Complexity:} Initially, neural networks were limited in size and complexity due to the computational constraints of the time. Early networks often consisted of only a few layers and a limited number of neurons, which constrained their learning capacity and applicability to complex tasks. However, as computational resources expanded, so too did the size and depth of these models. For instance, models like AlexNet and VGG in the early 2010s marked the beginning of what would be a rapid expansion in network depth and complexity, featuring layers deep into the double digits.

\textbf{Advancements in Hardware and Algorithms:} The advent of GPUs and improvements in distributed computing have significantly reduced the time required to train large neural networks. Simultaneously, advancements in optimization algorithms, such as Adam and RMSprop, have improved the efficiency of training deep networks. These technical advancements have facilitated the development of models such as the Transformer, which underpins modern NLP systems like GPT and BERT. These models not only have hundreds of layers but also millions to billions of parameters.

\textbf{Implications of Larger Models:} The shift towards larger models has resulted in substantial improvements in tasks such as image recognition, natural language processing, and complex decision-making processes. For example, larger models have led to breakthroughs in machine translation and autonomous vehicle technology. However, the trend towards larger networks also presents new challenges, including increased computational cost, energy consumption, and the need for more sophisticated techniques to combat overfitting.

\textbf{Future Prospects:} As the trend towards larger models continues, the field of machine learning is likely to witness even more sophisticated architectures. This progression suggests a future where neural networks could approach and even surpass human-like capabilities in certain tasks. However, this potential also necessitates innovations in model efficiency, training techniques, and hardware design to make the training and deployment of such models sustainable.

\subsubsection{The Necessity and Impact of Model Size Growth}
\textbf{The Necessity and Impact of Model Size Growth:} The necessity for growth in neural network model size stems primarily from the increasing complexity of tasks that modern AI systems are expected to perform. As the ambition to develop systems that can mimic human-level understanding and decision-making grows, so too does the need for models with greater capacity. Larger models, with their enhanced ability to model complex patterns and relationships, are pivotal in achieving higher levels of accuracy in tasks ranging from natural language understanding to complex image recognition.

\textbf{Impacts on Performance and Efficiency:} Larger neural network models have consistently set new benchmarks in AI performance. For instance, in natural language processing (NLP), models like OpenAI's GPT-3 have demonstrated remarkable linguistic understanding and generation capabilities, directly correlating their performance improvements to their vast number of parameters. Similarly, in image processing, larger Convolutional Neural Networks (CNNs) have achieved unprecedented accuracies in image classification challenges.

\textbf{Computational Challenges and Solutions:} However, the growth in model size is not without its challenges. The primary concern is the exponential increase in computational resources and energy required for training such large models. This has prompted significant research into more efficient training algorithms, pruning techniques, and specialized hardware accelerations like GPUs and TPUs, which are designed to handle extensive computational loads more efficiently.

\textbf{Economic and Environmental Considerations:} Moreover, the economic and environmental impact of training and deploying large-scale models cannot be overlooked. The financial cost associated with accessing the necessary computational power can be prohibitive, limiting the accessibility of cutting-edge AI technology to well-funded organizations. Environmentally, the carbon footprint associated with training and maintaining large models is substantial, prompting a push towards developing more energy-efficient computing techniques.

\textbf{Balancing Scale with Sustainability:} Moving forward, the challenge will be to balance the undeniable benefits of larger neural network models with the practical limitations they impose. Innovations in model design, such as the development of sparse networks and federated learning, offer promising avenues for maintaining model efficacy while mitigating computational and environmental costs. The future of neural networks, therefore, lies not just in scaling up, but in scaling smartly—enhancing model efficiency without compromising on their transformative potential.

\section{Objectives, Importance, and Fundamental Methods of Model Quantization}
\label{SEC:FundMethods}
\subsection{Fundamental Approaches}

\begin{figure*}[htbp]
    \centering
    \includegraphics[width=\linewidth]{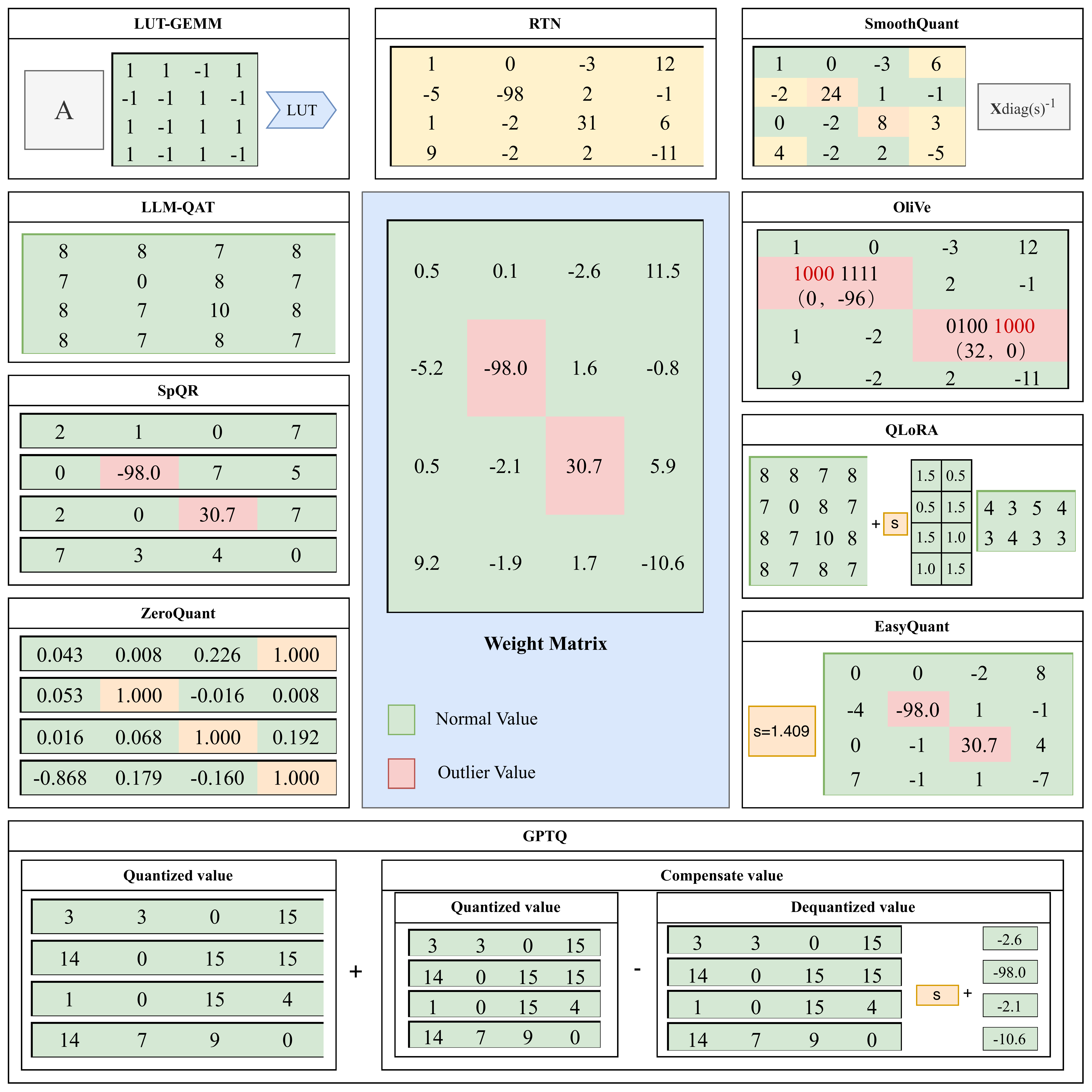}
    \caption{Comparison of Different Algorithms for Quantizing Weight Matrices. Some algorithms, such as RTN and LLM-QAT\cite{liu2024spinquantllmquantizationlearned}, directly quantize the weight matrix. Others, like SmoothQuant\cite{xiao2023smoothquant}, SpQR\cite{dettmers2023spqr}, OliVe\cite{guo2023olive}, and EasyQuant\cite{Tang2024EasyQuantAE}, process outliers separately from normal values. Algorithms like GPTQ\cite{frantar2022gptq} and QLoRa\cite{dettmers2024qlora} use matrix operation properties to preserve outliers during computations. Additionally, ZeroQuant\cite{yao2022zeroquant}, SpQR\cite{dettmers2023spqr}, and GPTQ\cite{frantar2022gptq} address fine-grained quantization issues.}
    \label{Comparison of Different Algorithms for Quantizing Weight Matrices}
\end{figure*}

\begin{table*}[ht]
    \centering
    \begin{tabular}{c|c}
    \toprule
        Model & Core Algorithm   \\ \midrule
        LLM-QAT\cite{liu2024spinquantllmquantizationlearned} & Data-free quantization-aware training  \\ 
        PEQA(L4Q)\cite{jeon2024l4q} & $Y = (W_0 + \alpha BA)X$  \\ 
        QLORA\cite{dettmers2024qlora} & NF4 Quantization + Double Quantization (DQ) + LoRA  \\ 
        LUT-GEMM\cite{park2022lut} & Pre-calculation+Look up table\\ 
        ZeroQuant\cite{yao2022zeroquant} & Fine-grained+LKD+Mixed quantizaiton/dequantization\\ 
        SmoothQuant\cite{xiao2023smoothquant} & Outliers Transfer:$Y = (Xdiag(s)^{-1}) (diag(s)W)$  \\ 
        SpQR\cite{dettmers2023spqr} & Sparse Quantization + Two-layer quantization  \\ 
        OliVe\cite{guo2023olive} & Sacrificing normal values for outliers   \\ 
        GPTQ\cite{frantar2022gptq} & MinMax quantization of approximate second-order information + adaptive batch update + Cholesky decomposition   \\ 
        AWQ\cite{lin2024awq} & Look for key weights by observing activations \\ 
        ACIQ\cite{banner2018aciq} & Clipping;Per-channel Bit Allocation \\ 
        LowbitQ\cite{Choukroun2019LowbitQO} & mutiple tensor  \\ 
        DFQ\cite{Nagel2019DataFreeQT} & equalizing weight ranges  \\ 
        PWLQ\cite{fang2020post} & Partition quantization  \\ 
        SPARQ\cite{Shomron2021PostTrainingSQ} & Select the most important bit as the quantized value  \\ 
        Easyquant\cite{Tang2024EasyQuantAE} & Isolate outliers for weight   \\ 
        BRECQ\cite{Li2021BRECQPT} & Block reconstruction  \\ 
        PTQD\cite{He2023PTQDAP} & quantization noise segmentation \\ 
        Zeroq\cite{Cai2020ZeroQAN} &distill data + Pareto frontier\\
        \bottomrule
    \end{tabular}
    \caption{Core Algorithm of Different Algorithms for Quantizing Large-Scale Models}
    \label{Core Algorithm}
\end{table*}

\begin{table*}[ht]
    \centering
    \begin{tabular}{c|ccc}
        \toprule
       \multirow{2}{*}{Model} & \multicolumn{3}{c}{Weight}  \\ 
              & Feature&Consider Outliers&Consider Importance \\
        \midrule
        LLM-QAT\cite{liu2024spinquantllmquantizationlearned} & Symmetrical MinMax quantization & ~ & \ding{51} \\ 
        PEQA(L4Q)\cite{jeon2024l4q} & LoRA & ~ & ~ \\ 
        QLORA\cite{dettmers2024qlora} & 4-bit NormalFloat (NF4) quantization & \ding{51} & \ding{51} \\ 
        LUT-GEMM\cite{park2022lut} & BCQ & ~ & ~ \\ 
        ZeroQuant\cite{yao2022zeroquant} & fine-grained+Group-wise & ~ & ~ \\ 
        SmoothQuant\cite{xiao2023smoothquant} & diag(s)W & \ding{51} & \ding{51} \\ 
        SpQR\cite{dettmers2023spqr} & Low bit width quantization + high precision 16-bit weight storage & \ding{51} & \ding{51} \\ 
        OliVe\cite{guo2023olive} & OVP+abfloat & \ding{51} & \ding{51} \\ 
        GPTQ\cite{frantar2022gptq} & MinMax quantization of approximate second-order information & \ding{51} & \ding{51} \\ 
        AWQ\cite{lin2024awq} & Determine the key parts by activating the values & ~ & ~ \\ 
        ACIQ\cite{banner2018aciq} & Per-channel bit allocation & ~ & ~ \\ 
        LowbitQ\cite{Choukroun2019LowbitQO} & kernel-wisely & ~ & ~ \\ 
        DFQ\cite{Nagel2019DataFreeQT} & equalizing ranges & ~ & ~ \\ 
        PWLQ\cite{fang2020post} & divide range into two regions & ~ & ~ \\ 
        Easyquant\cite{Tang2024EasyQuantAE} & leaving outliers unchanged & \ding{51} & ~ \\ 
        BRECQ\cite{Li2021BRECQPT} & Hessian matrix & ~ & ~ \\ 
        \bottomrule
    \end{tabular}
    \caption{Comparison of Matrix Quantization Across Different Large Model Quantization Algorithms. It displays the features and considerations of various algorithms in matrix quantization. The "\ding{51}" indicates that the respective algorithm considers outliers or the importance of weights during the quantization process.}
    \label{comparison1}
\end{table*}

\begin{table*}[ht]
    \centering
    \begin{tabular}{c|ccc}
        \toprule
       \multirow{2}{*}{Model} & \multicolumn{3}{c}{Activation}  \\ 
              & Feature&Consider Outliers&Consider Importance \\
        \midrule
        LLM-QAT\cite{liu2024spinquantllmquantizationlearned} & Activation quantization of each token & \ding{51} & \ding{51} \\ 
        QLORA\cite{dettmers2024qlora} & Brain Floating Point 16 (BFloat16) & \ding{51} & \ding{51} \\ 
        LUT-GEMM\cite{park2022lut} & Full precision & ~ & ~ \\ 
        ZeroQuant\cite{yao2022zeroquant} & Fine-grained+Token-wise & ~ & ~ \\ 
        SmoothQuant\cite{xiao2023smoothquant} & $Xdiag(s)^{-1}$ & ~ & ~ \\ 
        AWQ\cite{lin2024awq} & determine critical weights & \ding{51} & \ding{51} \\ 
        ACIQ\cite{banner2018aciq} & clip the range & ~ & \ding{51} \\ 
        LowbitQ\cite{Choukroun2019LowbitQO} & quantizing the residual & ~ & ~ \\ 
        DFQ\cite{Nagel2019DataFreeQT} & absorbs high biases & ~ & ~ \\
        PWLQ\cite{fang2020post} & same as weight & ~ & ~ \\
        Easyquant\cite{Tang2024EasyQuantAE} & 0: weight only & ~ & ~ \\ 
        BRECQ\cite{Li2021BRECQPT} & same as weight & ~ & ~ \\ 
        \bottomrule
    \end{tabular}
    \caption{Comparison of Activation Quantization Across Different Large Model Quantization Algorithms. It displays the features and considerations of various algorithms in Activation quantization. The "\ding{51}" indicates that the respective algorithm considers outliers or the importance of activation during the quantization process.}
    \label{comparison2}
\end{table*}

\begin{table*}[ht]
    \centering
    \begin{tabular}{c|cc|c|ccccc}
    \toprule
        Model & Memory Aligned & Trained & Knowledge Distillation Feature 
 & Bias Correction & Calibration Set & Mixed-precision & PTQ & QAT \\ \midrule
        LLM-QAT\cite{liu2024spinquantllmquantizationlearned} & ~ & \ding{51} & Logits Distillation & ~ & \ding{51} & ~ & ~ & \ding{51} \\
        PEQA(L4Q)\cite{jeon2024l4q} & ~ & \ding{51} & ~ & \ding{51} & \ding{51} & ~ & ~ & \ding{51} \\ 
        QLORA\cite{dettmers2024qlora} & \ding{51} & \ding{51} & ~ & ~ & ~ & ~ & ~ & \ding{51} \\
        LUT-GEMM\cite{park2022lut} & ~ & ~ & ~ & \ding{51} & ~ & \ding{51} & \ding{51} & ~ \\ 
        ZeroQuant\cite{yao2022zeroquant} & ~ & \ding{51} & LKD & ~ & ~ & ~ & \ding{51} & ~ \\
        SmoothQuant\cite{xiao2023smoothquant} & ~ & ~ & ~ & ~ & ~ & ~ & \ding{51} & ~ \\ 
        SpQR\cite{dettmers2023spqr} & \ding{51} & ~ & ~ & ~ & \ding{51} & ~ & \ding{51} & ~ \\
        OliVe\cite{guo2023olive} & \ding{51} & ~ & ~ & \ding{51} & ~ & ~ & \ding{51} & ~ \\
        GPTQ\cite{frantar2022gptq} & \ding{51} & ~ & ~ & ~ & \ding{51} & ~ & \ding{51} & ~ \\ 
        AWQ\cite{lin2024awq} & ~ & ~ & ~ & \ding{51} & ~ & \ding{51} & \ding{51} & ~ \\ 
        ACIQ\cite{banner2018aciq} & ~ & ~ & ~ & \ding{51} & ~ & \ding{51} & \ding{51} & ~ \\ 
        LowbitQ\cite{Choukroun2019LowbitQO} & ~ & \ding{51} & ~ & ~ & \ding{51} & ~ & \ding{51} & ~ \\ 
        DFQ\cite{Nagel2019DataFreeQT} & ~ & ~ & ~ & \ding{51} & ~ & ~ & \ding{51} & ~ \\ 
        PWLQ\cite{fang2020post} & ~ & ~ & ~ & \ding{51} & ~ & ~ & \ding{51} & ~ \\ 
        SPARQ\cite{Shomron2021PostTrainingSQ} & ~ & ~ & ~ & ~ & ~ & ~ & \ding{51} & ~ \\ 
        Easyquant\cite{Tang2024EasyQuantAE} & ~ & ~ & ~ & ~ & ~ & ~ & \ding{51} & ~ \\ 
        BRECQ\cite{Li2021BRECQPT} & ~ & ~ & ~ & ~ & ~ & \ding{51} & \ding{51} & ~ \\ 
        PTQD\cite{He2023PTQDAP} & ~ & ~ & ~ & \ding{51} & \ding{51} & \ding{51} & \ding{51} & ~ \\ 
        Zeroq\cite{Cai2020ZeroQAN} & ~ & ~ & ~ & ~ & ~ & \ding{51} & \ding{51} & ~ \\ \bottomrule
    \end{tabular}
    \caption{Comparison of Different Algorithms for Quantizing Large-Scale Models.  The "\ding{51}" symbol indicates that the specified feature or attribute is implemented or considered by the algorithm. This symbol helps to quickly identify which algorithms include certain functionalities, such as training, use of calibration sets, and implementation of quantization-aware training (QAT), among others. }
    \label{comparison3}
\end{table*}

\subsubsection{LLM-QAT}

LLM-QAT is an advanced method for Quantization-Aware Training (QAT) specifically designed for LLMs. Traditional post-training quantization methods have shown effectiveness up to 8-bits but struggle at lower bit precision levels. LLM-QAT leverages a data-free distillation technique that generates training data using the pre-trained model itself. This method helps in preserving the original output distribution and allows the quantization of weights, activations, and the key-value (KV) cache. The process aims to improve the efficiency and performance of LLMs even at quantization levels as low as 4-bits.

In detail, based on the observation, Symmetric MinMax quantization is first used to retain LLMs' outliers and maintain the performance of the model, which is defined as:

\[ 
X^i_Q = \alpha \left\lfloor \frac{X^i_R}{\alpha} \right\rfloor, \quad \alpha = \frac{\max(|X_R|)}{2^{N-1} - 1}, 
\]

where \(X_Q\) represents the quantized values, \(X_R\) represents the real (full-precision) values, and \(\alpha\) is the scaling factor. For weights, per-channel quantization is used, and for activations and the KV cache, per-token quantization is applied.

Second, LLM-QAT uses a student-teacher model framework to ensure that the quantized model retains the performance of the full-precision model. Specifically, the student network, which is the lower-precision version of the model, is guided by the full-precision teacher network, the original pre-trained model. This guidance is provided through cross-entropy-based logits distillation:

\[ L_{CE} = - \frac{1}{n} \sum_{c} \sum_{i=1}^{n} p^T_c(X_i) \log(p^S_c(X_i)) \]

Here, \(i\) represents the \(i\)-th sample in the batch, \(c\) denotes the number of classes (vocabulary size), and \(T\) and \(S\) are the teacher and student networks, respectively. 

Third, next token data generation from the pre-trained model is proposed for synthesizing the distribution of pre-training data. Data is generated by initiating with a random start token and generating subsequent tokens iteratively until the end of the sentence or maximum length is reached. And to ensure the generated data is diverse and accurate, LLM-QAT introduces a a hybrid approach where the first few tokens are deterministically selected with top-1 strategy and the remaining tokens are stochastically sampled from the pre-trained model's SoftMax output distribution. 

Lastly, The generated data is then used as input for fine-tuning the quantized model, where the teacher model's predictions serve as labels to guide the training.

Experimental results show that LLM-QAT significantly outperforms traditional PTQ methods at lower bit precisions. For example, in the 8-8-4 setting, the 30B LLM-QAT model achieves an average zero-shot accuracy of 69.7, compared to 50.7 with SmoothQuant, demonstrating its robustness in maintaining accuracy. In the 4-8-4 setting, where both weights and the KV cache are quantized to 4 bits, LLM-QAT achieves 69.9, only 1.5 points behind the full-precision model, while all PTQ methods perform poorly, highlighting LLM-QAT’s superior quantization capabilities. 

Additionally, in the 4-8-8 setting, LLM-QAT outperforms the best PTQ method (RTN) by 1.4 points. These results are consistent across different model sizes, with an 8-8-8 30B quantized model surpassing a 13B full-precision model in performance, and a 4-8-4 LLM-QAT 30B model outperforming an 8-bit LLaMA-13B. These findings underscore LLM-QAT's ability to maintain high performance with reduced computational costs and memory usage, offering a better efficiency-accuracy tradeoff.

\subsubsection{PEQA}

To address the increasing memory and computational costs in large-scale NLP models, researchers Hyesung Jeon, Yulhwa Kim, and Jae-Joon Kim from Seoul National University and Sungkyunkwan University proposed the \textbf{L}ow-rank adaptive \textbf{L}earning quantization algorithm geared towards \textbf{LL}Ms  (L4Q)\cite{jeon2024l4q} . L4Q combines quantization and parameter-efficient fine-tuning (PEFT) to overcome the limitations of traditional methods. While Post-Training Quantization (PTQ) is efficient but error-prone, and Quantization-Aware Training (QAT) is accurate but resource-intensive, L4Q integrates QAT with PEFT to achieve both high precision and low memory usage, making it ideal for resource-constrained environments.

L4Q achieves the integration of quantization and fine-tuning through the following steps:

\begin{enumerate}
    \item Merging Weights and LoRA Parameters:
   The pre-trained weights \( W_{FP} \) and LoRA parameters \( A \) and \( B \) are merged into a new weight matrix:
     \[
     W' = W_{FP} + \alpha BA
     \]

    \item Quantizing the New Weight Matrix:
    The merged weight matrix \( W' \) is quantized to generate quantized weights \( W_q \):
     \[
     W_q = R(\text{clamp}(\frac{W' - b}{s}, Q_n, Q_p)) \times s + b
     \]
     \( R \) represents the rounding function, and \(\text{clamp}\) function limits the values within the quantization range.

    \item Optimizing Quantization Parameters:
    L4Q uses the LSQ method to update the quantization parameters \( s \) and \( b \):
     \[
     \frac{\partial L}{\partial s} = -w + \tilde{w}, \quad \text{if } Q_n \leq w \leq Q_p
     \]
     \[
     \frac{\partial L}{\partial b} = 1, \quad \text{if } w < Q_n \text{ or } w > Q_p
     \]

    \item Gradient Calculation for LoRA Parameters:
   The quantized gradients \(\frac{\partial L}{\partial W_q}\) are used to update the LoRA parameters \( A \) and \( B \):
     \[
     \frac{\partial L}{\partial A} = \alpha B^T \frac{\partial L}{\partial W_q}
     \]
     \[
     \frac{\partial L}{\partial B} = \alpha \frac{\partial L}{\partial W_q} A^T
     \]
\end{enumerate}

Experimental results show that the L4Q method significantly improves performance across various tasks. In the Commonsense QA benchmark, L4Q achieves higher model accuracy in most configurations, especially by $2\%$ in 3-bit quantization scenarios. In the MMLU benchmark tests, L4Q outperforms QLoRA and QA-LoRA in zero-shot (0-shot) and few-shot (5-shot) tasks, with accuracy improvements of approximately $1.5\%$ and $2\%$ respectively. These advantages make L4Q highly valuable in the field of large model quantization.

\begin{figure*}
    \centering
    \includegraphics[width=\linewidth]{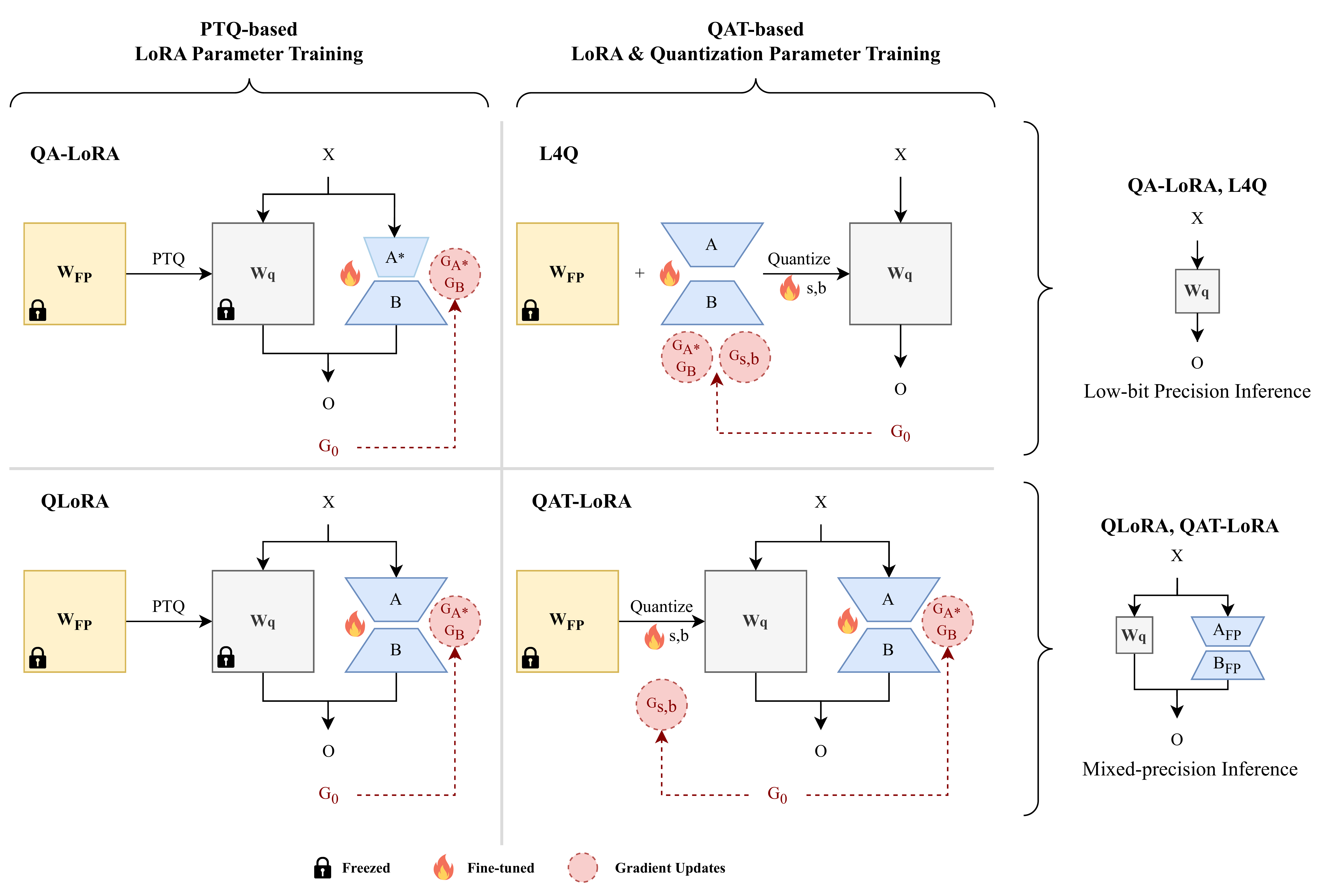}
    \caption{L4Q Comparison from \cite{jeon2024l4q}}
    \label{fig:L4Q}
\end{figure*}

\subsubsection{QLORA}

\subsubsection{LUT-GEMM}

To address the increasing memory and computational costs in large-scale NLP models, researchers Gunho Park and Baeseong Park from Pohang University of Science and Technology and NAVER Cloud proposed the \textbf{L}ookup \textbf{T}able-based \textbf{G}EMM (LUT-GEMM) algorithm\cite{park2022lut}. LUT-GEMM enhances inference efficiency by quantizing weights while maintaining full precision for activations, eliminating the dequantization step.

LUT-GEMM achieves quantized matrix multiplication through the following steps:

\begin{enumerate}
    \item Constructing Lookup Tables:
   For a binary matrix \(B \in \{-1, +1\}^{m \times n}\) and an activation vector \(x \in \mathbb{R}^n\), all possible combinations of activation values and binary patterns are precomputed and stored in a lookup table (LUT). This can be expressed as:
   \[
   w \approx \sum_{i=1}^{q} \alpha_i b_i
   \]
   where \(\alpha_i \in \mathbb{R}^+\) are scaling factors and \(b_i \in \{-1, +1\}^n\) are binary vectors.

    \item Lookup Table Retrieval:
   Using the LUT, precomputed partial dot products are retrieved by indexing directly, avoiding redundant computations. For matrix multiplication \(Bx^T\), the LUT retrieval operation replaces the original calculations:
   \[
   Bx^T = \text{LUT}[B, x]
   \]

    \item Reducing Computational Complexity:
   By utilizing the LUT, LUT-GEMM reduces the complexity of computations. Assuming a q-bit quantized binary matrix \(B_i\) multiplied by an input vector \(x\), the computational complexity is:
   \[
   C = O \left( m \cdot \frac{n}{\mu} \cdot q \right)
   \]
   where \(\mu\) is the length of the sub-vector.

    \item GPU Parallel Implementation:
   For implementation on GPUs, LUT-GEMM improves parallelism by assigning as many threads as possible to perform independent LUT accesses, with scaling factor operations not degrading thread performance. Each thread block (TB) shares LUTs, utilizing the high bandwidth of GPUs for fast matrix computations.
\end{enumerate}

Experimental results show that when applied to the OPT-175B model with 3-bit quantization, LUT-GEMM substantially accelerates token generation latency, achieving a remarkable 2.1× improvement on a single GPU compared to OPTQ, which relies on the costly dequantization process.The experimental results clearly demonstrate that LUT-GEMM achieves the lowest latency by directly using quantized weights and reducing computational complexity, significantly saving energy consumption.

\subsubsection{ZeroQuant}

With the increasing size of Transformer models like BERT and GPT, the computational and memory costs during inference have become a significant challenge, even for powerful GPUs. To address this, a research team at Microsoft proposed the ZeroQuant algorithm at NeurIPS 2022\cite{yao2022zeroquant}. ZeroQuant aims to efficiently compress large-scale Transformer models through post-training quantization (PTQ), eliminating the need for retraining.

The core of the ZeroQuant algorithm lies in its proposal of a fine-grained, hardware-friendly quantization strategy, combined with Layer-wise Knowledge Distillation (LKD), allowing the maintenance of high model accuracy even under extreme low-bit-width quantization (e.g., INT4).

\textbf{Fine-Grained Quantization Strategy}

The quantization strategy proposed by ZeroQuant consists of two parts: Group-wise Quantization and Token-wise Quantization.

- Group-wise Quantization: Traditional quantization methods typically apply uniform quantization to the entire weight matrix, which can lead to significant accuracy loss when applied to large-scale models. ZeroQuant mitigates this by dividing the weight matrix into multiple groups and quantizing each group separately, thereby reducing quantization errors and improving hardware efficiency.

  The quantization is expressed by the following formula:

  \[
  x_{\text{quantize}} = \text{round}\left(\text{clamp}\left(\frac{x}{S}, -2^{\text{bit}-1}, 2^{\text{bit}-1} - 1\right)\right)
  \]

  where \(x\) is the vector or matrix to be quantized, \(S\) is the scaling factor (usually the maximum absolute value), and \(\text{bit}\) is the quantization bit width.

- Token-wise Quantization: For the quantization of activations, ZeroQuant adopts a more refined strategy. Due to the significant variance in activation ranges across different tokens in Transformer models, a uniform quantization range often results in substantial performance degradation. Therefore, ZeroQuant proposes a token-wise quantization scheme, which dynamically calculates the quantization range for each token, minimizing quantization errors.

\textbf{Layer-wise Knowledge Distillation (LKD)}
The ZeroQuant algorithm introduces Layer-wise Knowledge Distillation (LKD) to address the accuracy issues associated with extreme low-bit-width quantization (e.g., INT4). Unlike traditional knowledge distillation methods, LKD does not require keeping full copies of both the teacher and student models. Instead, it quantizes and distills the model layer by layer, significantly reducing GPU memory requirements and enabling efficient quantization even without access to the original training data.

The distillation loss function for LKD is as follows:

\[
L_{\text{LKD}, k} = \text{MSE}\left(L_k \cdot L_{k-1} \cdot \dots \cdot L_1(X) - \hat{L}_k \cdot L_{k-1} \cdot \dots \cdot L_1(X)\right)
\]

where \(L_k\) represents the original model weights of the k-th layer, \(\hat{L}_k\) represents the quantized weights, \(X\) is the input data, and MSE denotes the mean square error.

ZeroQuant demonstrates remarkable efficiency in model quantization, reducing the precision of weights and activations to INT8 for both BERT and GPT-3 models without retraining, achieving up to 5.19x and 4.16x speedup compared to FP16 inference with minimal accuracy impact. When combined with Layer-wise Knowledge Distillation (LKD), ZeroQuant can further quantize fully-connected module weights to INT4, while keeping attention module weights and activations at INT8, leading to a 3x reduction in memory footprint compared to FP16 models. Additionally, ZeroQuant has been successfully applied to large open-source language models like GPT-J6B and GPT-NeoX20B, where the INT8 versions maintain similar accuracy to FP16 but with up to 5.2x greater efficiency.

\subsubsection{SmoothQuant}

In response to the growing memory and computational costs of large-scale NLP models, researchers Guangxuan Xiao, Ji Lin from the Massachusetts Institute of Technology (MIT), and Mickael Seznec, Hao Wu, Julien Demouth from NVIDIA proposed a post-training quantization (PTQ) method for large language models (LLMs) called \textbf{SmoothQuant}\cite{xiao2023smoothquant}. This method specifically addresses the issues of maintaining accuracy and hardware efficiency during quantization by implementing an INT8 quantization for both activations and weights through a training-free transformation, optimizing model execution efficiency and memory usage on hardware.

\textbf{SmoothQuant} centers around several key components:

\begin{enumerate}
    \item Quantization Transformation Formula:
   SmoothQuant reduces quantization difficulty by smoothing input activations \(X\) and adjusting weights \(W\). The core transformation formula is:
   \[
   Y = (X\text{diag}(s)^{-1}) \cdot (\text{diag}(s)W) = \hat{X}\hat{W}
   \]
   Here, \(s\) represents the smoothing factor for each channel, making the adjusted activations \(\hat{X}\) and weights \(\hat{W}\) easier to quantize.

    \item Selection of Smoothing Factors:
   The choice of smoothing factor \(s_j\) is aimed at maximizing quantization effectiveness and accuracy, calculated by:
   \[
   s_j = \frac{\max(|X_j|)^\alpha}{\max(|W_j|)^{1-\alpha}}
   \]
   \(\alpha\) is a hyperparameter between 0 and 1 that balances the quantization difficulty between activations and weights.

    \item Application to Transformer Blocks:
   Within Transformer models, SmoothQuant specifically applies scaling smoothing to the input activations of self-attention and feed-forward layers, and uses INT8 quantization for all linear layers involving both weights and activations.

\end{enumerate}

In practical applications, SmoothQuant demonstrated significant performance enhancements across various configurations and large models. For instance, in tests using the OPT-175B model, SmoothQuant achieved a 1.51x speed improvement and 1.96x memory savings almost without loss of accuracy. This method not only maintains the model's precision but also significantly enhances hardware efficiency, especially valuable in resource-constrained environments.

\subsubsection{SpQR}

\subsubsection{OliVe}

To address the increasing memory and computational costs in large-scale NLP models, researchers Cong Guo, Jiaming Tang, Weiming Hu, and others from Shanghai Jiao Tong University and Microsoft Research proposed the \textbf{Outlier-Victim Pair Quantization algorithm} (OliVe)\cite{guo2023olive} for large language models. OliVe employs a hardware-friendly method to handle outliers, significantly enhancing performance and energy efficiency while maintaining model accuracy in resource-constrained environments.

The OliVe algorithm achieves effective handling of outliers through the following steps:

\begin{enumerate}
    \item Pair-wise Analysis: OliVe first analyzes the tensor values in the model, classifying them into three types of pairs: normal-normal, outlier-normal, and outlier-outlier. The core of the algorithm is that for outlier-normal pairs, the normal values are set to zero (referred to as "victims"), freeing up space to handle the outliers.

\item Outlier Quantization: OliVe uses an adaptive biased float (abfloat) data type to quantize outliers. This method adds a suitable bias to adjust the range of floating-point values, ensuring they do not overlap with the range of normal values, thus maximizing the utilization of the numerical representation space. Specifically, the abfloat values are quantized using the formula:
   \[
   \text{quant}(e) = \text{sign} \times (1 \ll \text{mantissa} + \text{mantissa}) \ll (\text{exponent} + \text{bias})
   \]
   where \(\text{mantissa}\) denotes the width of the mantissa bits, \(\text{exponent}\) denotes the width of the exponent bits, and \(\text{bias}\) is the adaptive bias.

\item Hardware-Friendly Memory Alignment: A key feature of OliVe is memory alignment. By positioning the "victims" adjacent to the outliers, OliVe achieves efficient memory access with low hardware overhead. This design avoids the complexity of sparse indexing hardware and is compatible with the memory subsystems of existing accelerators such as GPUs and TPUs.

\end{enumerate}

In terms of experimental results, OliVe demonstrated significant improvements across various tasks and datasets. For instance, in the GLUE benchmark with the BERT model, the 4-bit Post-Training Quantization (PTQ) method of OliVe resulted in less than a $1\%$ drop in accuracy compared to the full-precision model, outperforming other 4-bit, 6-bit, and 8-bit PTQ and Quantization-Aware Training (QAT) methods. Additionally, when applied to large-scale language models like GPT2-XL, BLOOM-7B1, and OPT-6.7B, OliVe's 8-bit PTQ method nearly preserved the original model performance, highlighting its superior inference performance and energy efficiency.

\subsubsection{GPTQ}

\subsubsection{AWQ}

In the field of quantizing large language models (LLMs), traditional methods face significant challenges such as high training costs and notable accuracy loss at low bit rates. A research team from MIT, including Ji Lin, Jiaming Tang, Haotian Tang, Shang Yang, Wei-Ming Chen, Wei-Chen Wang, Guangxuan Xiao, Xingyu Dang, Chuang Gan, and Song Han, proposed an algorithm called Activation-aware Weight Quantization (AWQ)\cite{lin2024awq} to address these challenges. AWQ is a hardware-friendly low-bit weight-only quantization method based on the observation that weights are not equally important, and protecting only $1\%$ of salient weights can significantly reduce quantization error. AWQ protects salient weights by observing activations rather than weights themselves, requiring no backpropagation or reconstruction, thereby maintaining the generalization ability of LLMs across different domains and modalities without overfitting to the calibration set. 

The core idea of the AWQ algorithm is based on the observation that weights are not equally important, and protecting only $1\%$ of the significant weights can greatly reduce quantization error. The algorithm protects significant weights by observing activations rather than the weights themselves. The specific steps are as follows:

\begin{enumerate}
    \item Selecting Significant Weights: Significant weight channels are selected based on the activation distribution rather than the weight distribution. This is because weight channels associated with larger activation magnitudes process more important features.

    \item Weight Channel Scaling: A per-channel scaling method is designed to automatically search for the optimal scaling factor to minimize quantization error under full-weight quantization. The quantization function is defined as:
   \[
   Q(w) = \Delta \cdot \text{Round}\left(\frac{w}{\Delta}\right), \Delta = \frac{\text{max}(|w|)}{2^{N-1}}
   \]
   where \(N\) is the number of quantization bits, and \(\Delta\) is the quantization scale determined by the maximum value of the weights.

   To protect significant weights, quantize before scaling the weights:
   \[
   Q(w \cdot s) \left(\frac{x}{s}\right) = \Delta' \cdot \text{Round}\left(\frac{w \cdot s}{\Delta'}\right) \cdot \frac{x}{s}
   \]
   By analyzing the error, determine the scaling factor \(s\) to reduce the quantization error of significant weights, with the optimization objective:
   \[
   s^* = \arg \min_s \| Q(W \cdot \text{diag}(s))( \text{diag}(s)^{-1} \cdot X) - WX \|
   \]
\end{enumerate}

Experimental results show that AWQ outperforms existing methods on various language modeling and domain-specific benchmarks, especially in instruction-tuned and multi-modal models. The team also implemented TinyChat, an efficient and flexible inference framework for deploying 4-bit quantized LLMs on various edge platforms, achieving more than 3x speedup compared to the Huggingface FP16 implementation.

\subsection{The Significance of Quantization in Modern Deep Learning}

\section{Training with Quantization: Algorithms and Approaches}
\label{SEC:TraQuant}
\subsection{Quantized Neural Networks (QNNs)}

Quantized Neural Networks(QNNs) are variants of neural network that use quantized weights and/or activations instead of traditional full-precision (32-bit floating point) numbers. The motivation behind quantization   els with more and more weights, which are not suitable for deployment on the resource-constrained devices such as mobile phones, embedded system and IoT devices. And the goal of quantization is to reduce the computational cost and memory requirements of neural networks without performance degradation. 

\subsection{Training neural networks With Quantization}

When quantizing neural networks, there are three components could be quantized: weights, activations and gradients. The quantization of weights and activations could effectively reduce the size of models, and the quantization of gradients could reduce the resource and time cost when training QNNs. 

Before the advent of Large Language Models(LLMs), the QNNs had already attracted a lot of attention. Many researchers dived into the quantization of weights or activations, and proposed a lot of techniques to reduce the cost of deployments and accelerate the inference. However, due to the need of the fidelity of gradient computations during backward propagation, the training of QNNs is unlike the inference, which could work well on low-precision. There is a high probability that the models will not converge if training QNNs with low-precision gradients. Therefore, the most quantization techniques try to quantize weights and/or activations of well-trained neural networks rather than train a QNNs from scratch. 

BinaryConnect(BC)\cite{courbariaux2015binaryconnect} proposes a method in training DNNs with binary weights during propagations. This work aims to use 1-bit to present weights $w$, and gives two quantization method $Q_w$ to convert $w$ to 1-bit. One of that is deterministic method which is based on the sign function:
\begin{equation}\label{eq1}
    w_q=Q_w(w)=\left\{
        \begin{aligned} 
            +1 &, & if \ w \geq 0, \\
            -1 &, & otherwise
        \end{aligned}
    \right.
\end{equation}
This method make the weights greater than 0 to become 1 and the other become -1. And the other is stochastic method:
\begin{equation}\label{eq2}
    w_q=Q_w(w)=\left\{
        \begin{aligned} 
            +1 &,\ with\ probability\ p = \delta(w), \\
            -1 &,\ with\ probability\ 1 - p
        \end{aligned}
    \right.
\end{equation}
where $\delta$ is the $hard sigmoid$ function:
\begin{equation}\label{eq3}
\delta(x) = \mathrm{clip}(\frac{x + 1}{2}, 0, 1)
\end{equation}
which convert the weights greater than 1 to 1, the weights less than -1 to -1 and the weights between -1 and 1 to either -1 or 1 with a probability.

During backward propagation, the BC also adopt the gradient decent algorithm like full-precision neural networks. However the common quantization function is not differentiable, or the derivative value is 0 almost everywhere(e.g., sign function mentioned above) which will make the gradient vanish. Therefor, They use the straight-through estimator(STE) to obtain gradients. This technique has been proposed by Hinton et al. at 2012. The STE function is defined as follow:
\begin{equation}\label{eq4}
    \mathrm{STE}(x)=\mathrm{clip}(x, -1, 1) = min(1, max(x, -1))
\end{equation}

During the forward propagation, the BC us $w_b=Q_w(w_r)=sign(w_r)$ to binarize the weights and forward with binary weights. During backward propagation, the gradient of cost C respect $w_r$ as the follow equation:
\begin{equation}\label{eq5}
    \frac{\partial{C}}{\partial{w_r}}=\frac{\partial{C}}{\partial{w_b}}\frac{\partial{w_b}}{\partial{w_r}}
\end{equation}
And through STE to obtain an estimate for the gradient of $w_b$ respect $w_r$:
\begin{equation}\label{eq6}
    \frac{\partial{w_b}}{\partial{w_r}} = I_{\vert w_r \vert}
\end{equation}
where $I_{\vert w_r \vert}$ is the derivative of STE:
\begin{equation}\label{eq7}
    I_{\vert w_r \vert}=\left\{
    \begin{aligned}
        1 &, \ if\ \vert w_r \vert \leq 0, \\
        0 &, \ \mathrm{otherwise}
    \end{aligned}
    \right.
\end{equation}
The estimation could add noisy to weights as a form of regulation.
Above all, the BC can correctly use gradient decent algorithm to update the weights, and obtain $98.8\%$ accuracy on MNIST dataset.

BNN(Binary Neural Networks)\cite{courbariaux2016binarized} , which based on the BinaryConnect and quantizes the activations further. Besides that, considering that batch normalization can help the networks avoid the problem of exploding and vanishing gradients, accelerate the training, and ensure that BNN can converge. But the batch normalization operation cannot work efficiently without floating point unit. In the light of that, BNN simplifies batch normalization to a shift-based variation.

XNOR-Net\cite{rastegari2016xnor} implements the convolution operation using XNOR operations and bit-count operations as the follow equation:
\begin{equation}\label{eq8}
    \bm{x}\cdot\bm{y}=\mathrm{bitcount}(\mathrm{and}(\bm{x}, \bm{y})), x_i, y_i \in \{0, 1\}
\end{equation}

DoReFa-Net\cite{zhou2016dorefa}, which has first quantized gradients to low-bitwidth(less than 8) floating-point numbers with discrete states in backward propagation. But when updating the weights, DoReFa-Net also use the full-precision weights and gradients.

WAGE\cite{wu2018training} realize the most work still keep high precision and computational complexity during training.(e.g. BC and BNN maintain the full-precision gradients and weights during backpropagation.) They develop a new method termed as WAGE to quantize both training and inference. In WAGE the weights, activations, gradients, and errors among layers are shifted and linearly constrained to low-bitwidth integers. It is the first work that uses the discrete gradients to update the quantized weights. WAGE uses shift-based linear mapping and stochastic rounding technique for quantization. In WAGE quantization, the continuous and unbounded values are discretized by uniform distance $\delta$:
\begin{equation}\label{eq9}
    \delta(k) = 2^{1-k},\ k \in \mathbb{N}_+
\end{equation}
and the use linear mapping function as basic quantization function to convert the floating-point number $x$ to $k$-bitwith singed integer. The basic function is defined as the follow:
\begin{equation}\label{eq10}
    Q(x,k) = \mathrm{clip}\{\delta(k)\cdot\mathrm{round}[\frac{x}{\delta(k)}],-1+\delta(k), 1-\delta(k)\}
\end{equation}
where round function approximate a value to the nearest discrete value.

Before the linear mapping function applied, they shift values distribution to an appropriate order of magnitude. The shift function is defined as the follow:
\begin{equation}\label{eq11}
    \mathrm{shift}(x) = 2^{\mathrm{round}(log_2^x)}
\end{equation}

Weights are quantized by the linear mapping function above and activations are quantized by the same function after a shift-based batch normalization.

During backpropagation, WAGE first scales gradients $g$ to the minimum step number.(for $k$-bitsinteger the minimum step is $\pm1$, for floating-point the minimum step is $\pm\delta(k)$)and keeps its direction. Then bring in shift-based learning rate $\eta$.

\begin{equation}\label{eq12}
    g_s = \eta \cdot g / \mathrm{shift}(\mathrm{max}(|g|))
\end{equation}

where $\eta$ is an power of 2 integer. It is used to control update step sizes of weights. To substitute accumulation of small gradients, WAGE separates $g_s$ into integer part and decimal part, and use $\mathrm{bernoulli}$ function to sample the decimal part to either $0$ or $1$. 
Above all, weights can be updated properly:
\begin{equation}\label{eq13}
    \delta W = Q_G(g) = \delta(k) \cdot \mathrm{sgn}(g_s) \cdot \{\lfloor \vert g_s \vert \rfloor  + \mathrm{bernoulli}(\vert g_s \vert - \lfloor \vert g_s \vert \rfloor) \}
\end{equation}

Besides STE, there is another way to solve the difficulty in propagating gradients which raises from non-differential quantization function.  Zhuang et al.proposed a solution by training the low-precision network with a full-precision auxiliary module, creating additional full-precision routes to optimize the low-precision model.\cite{zhuang2020training}

Besides training the QNNs via gradient decent algorithm, Peng et al. use the Evolutionary Algorithms(EAs) to search for the optimal low-bits weights of QNNs.\cite{peng2022training} They formulate the quantization of neural networks as a large-scale discrete optimization problem.

\subsection{Quantization after the advent of LLMs}

With the advent of LLMs, the excellent performance and the huge resource cost and memory requirement create a huge demand for low cost deployment and fast inference. Much and much attention has been attracted in the quantiaziton of LLMs. But to the best of our knowledge, there doesn't exist a work that pre-train a quantized LLM so far. Considering that training a LLM is very expensive, the pre-train of LLMs has been monopolized by large corporations. There is a very urgent that propose techniques to effectively reduce the resource and time cost in LLMs pre-train. In our opinion, the difficulty of Pre-training Quantized LLMs lies in the following points: 

\begin{itemize}
\item LLMs have too many parameters. The more parameters there are, the harder it becomes to predict the overall impact on training after quantizing.

\item LLMs are deeper than traditional DNNs. As the depth increases, the problem of exploding and vanishing gradients worsens.
\end{itemize}

\section{Inference with Quantization: Algorithms and Approaches}
\label{SEC:InfQuant}

\subsection{Knowledge Distillation (KD)}

Knowledge Distillation (KD) is a critical technique for model compression, enabling the transfer of knowledge from a large, well-trained teacher model to a smaller, more efficient student model. This process is particularly valuable in Quantization-Aware Training (QAT), where models are trained to maintain high performance despite being quantized to lower precision to reduce memory and computational costs. The primary motivation behind using KD in QAT is to enhance the performance of quantized models. KD helps mitigate the accuracy loss typically associated with model quantization by transferring the knowledge encapsulated in the teacher's full-precision model to the student’s lower-precision model. This process is particularly beneficial for resource-constrained environments where computational efficiency is paramount.

A. Key Techniques in KD for QAT

\begin{enumerate}
    \item Layer-wise Distillation:
    \begin{itemize}
        \item Method: Applying KD at each layer of the model, aligning the student’s intermediate representations with those of the teacher. This approach helps in capturing the hierarchical features learned by the teacher model.
        \item Benefits: This technique has been shown to significantly improve the performance of quantized models by maintaining the structural and 
        functional integrity of the model across all layers.
    \end{itemize}
    \item Attention Mechanism Distillation:
    \begin{itemize}
        \item Method: Transferring attention maps from the teacher to the student model. This ensures that the student model focuses on similar regions as the teacher, preserving the interpretability and effectiveness of the attention mechanism.
        \item Advantages: Enhances the student model’s performance by ensuring that the critical attention mechanisms learned by the teacher are retained in the quantized student model.
    \end{itemize}
    \item Logit-based Distillation:
    \begin{itemize}
        \item Method: Aligning the logits (pre-softmax outputs) of the student model with those of the teacher. This ensures that the probability distributions produced by the student are similar to those of the teacher, aiding in better generalization.
        \item Benefits: This method is straightforward and effective, providing a strong baseline for performance improvement in quantized models.
    \end{itemize}
\end{enumerate}
B. Advanced KD Techniques for QAT

\begin{enumerate}
    \item Quantization-aware Knowledge Distillation (QKD):
    \begin{itemize}
        \item Phases:QKD typically involves three phases - self-studying, co-studying, and tutoring. These phases help in progressively transferring knowledge and adapting the student model to quantized constraints.
        \item Technical Details: QKD uses a trainable uniform quantization scheme for weights and activations, along with gradient approximation techniques such as the straight-through estimator (STE) to handle the non-differentiable nature of quantizers.
    \end{itemize}
    
    \item Self-Supervised Quantization-Aware Knowledge Distillation (SQAKD):
    \begin{itemize}
        \item Methodology:SQAKD integrates QAT and KD by framing the problem as a constrained optimization task. It utilizes self-supervised learning techniques to minimize both discretization error and prediction discrepancy, ensuring precise quantization without sacrificing accuracy.
        \item Performance: SQAKD has demonstrated significant performance improvements across various architectures and datasets, showing its robustness and effectiveness in enhancing the accuracy of low-bit quantized models.
    \end{itemize}
\end{enumerate}
The effectiveness of KD methods in QAT has been evaluated extensively on various benchmark datasets, including CIFAR-10, CIFAR-100, and ImageNet. These evaluations show that KD significantly enhances the performance of student models, even with aggressive quantization (e.g., sub-2-bit precision). Metrics such as cover length ratio and ranking loss have been introduced to quantitatively assess the effectiveness of KD losses, further validating the improvements achieved through these techniques.

\begin{figure*}
    \centering
    \includegraphics[width=1\linewidth]{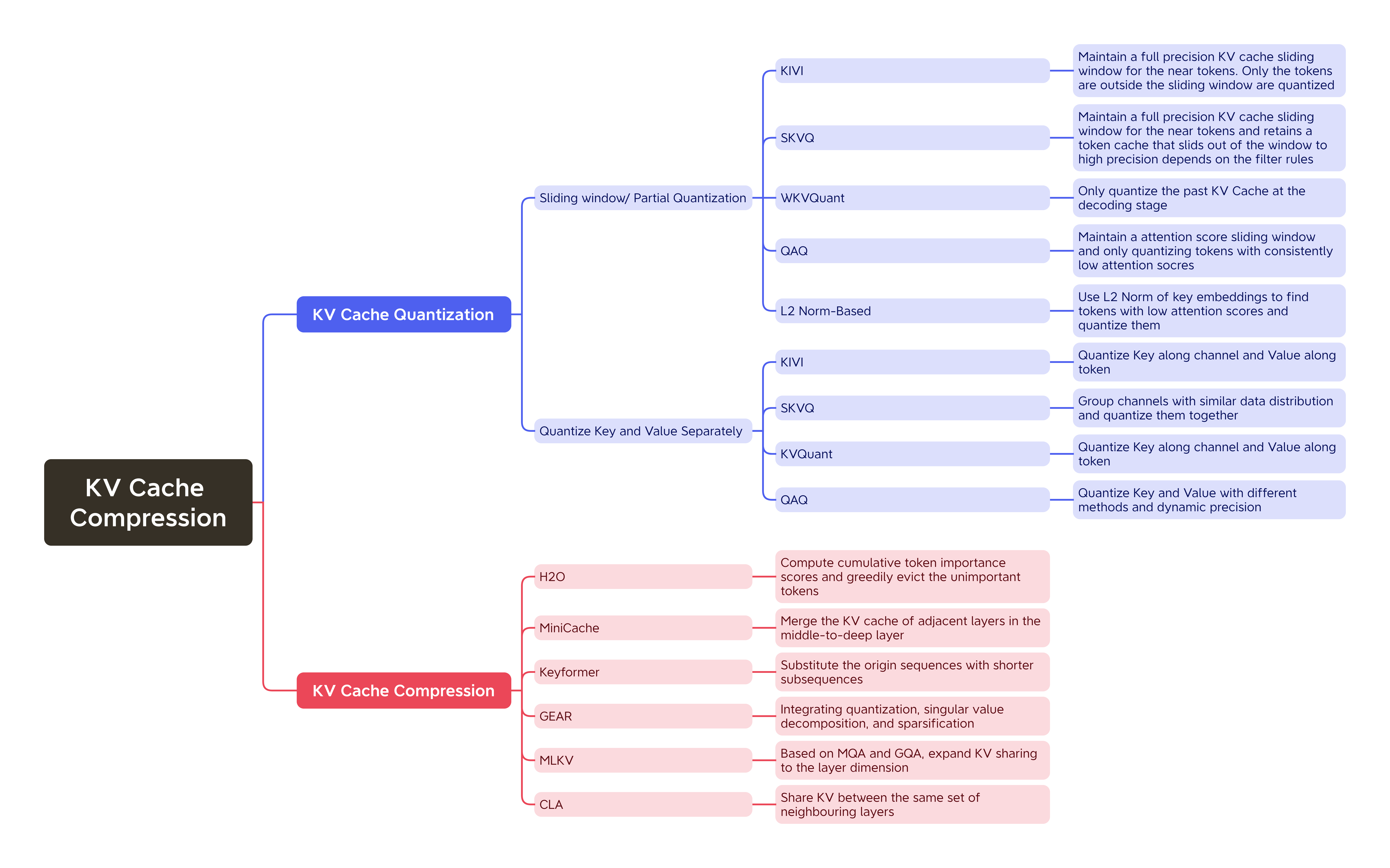}
    \caption{Summary of KV Cache Compression}
    \label{fig:enter-label}
\end{figure*}

\subsection{Key-Value cache(KV cache) compression}

In large-scale neural network computation, the concept of Key-Value Cache (KV Cache) is mainly related to the self-attention mechanism commonly used in the Transformer architecture.The purpose of KV Cache is to reduce the computational complexity, improve the efficiency, and conserve the computational resources, and it is widely used in large-scale models.

In Transformer, the operation of the self-attention mechanism involves the computation of query (Q), key (K) and value (V) matrices\cite{vaswani2017attention}. These matrices are used to compute the attention distribution to balance the input information at different locations. During inference, the Q matrix is usually derived from the model input, which makes the Q matrix different for each input instance. In contrast, the K matrix and the V matrix are computed from the output of the encoder and are relatively stable.

The key idea behind the KV cache is that due to the relative stability of the $K$ and $V$ matrices, they can be cached at different time steps. This means that for the same input, there is no need to recalculate the K and V matrices and therefore they can be reused.The KV cache can store the result of multiplying each token with the $W_K $ and $W_V $ parameter matrices. In the converter architecture, each token is generated based on the result of the previous token. the KV cache caches the K and V matrices, thereby reducing computation time. Without the KV cache, recomputing the product of $W_K $ and $W_V $ for all tokens each time a new token is generated would be computationally intensive. Therefore, caching these results is known as KV caching, which can improve the speed of reasoning.

To avoid recalculating attention keys and values, LLMs store previously computed keys and values, which leads to significant memory consumption as batch sizes and sequence lengths increase. For instance, In the experiment of KVQuant\cite{hooper2024kvquant}, they find that with the LLaMA-7B model, the KV cache consumes only 2\% of memory during inference with a sequence length of 512. However, this usage skyrockets to 84\% when the sequence length increases to 128k.

In the field of KV cache compression methods, two hardware optimisation methods: ‘DeepSpeed Inference’ \cite{aminabadi2022deepspeed}and ‘FlexGen’\cite{sheng2023flexgen}.DeepSpeed Inference, together with ZeRO Inference, introduces a multi-GPU inference solution that leverages heterogeneous memory (including GPU memory, DRAM, and NVMe) to meet the demands of large-scale models and improve inference performance. Instead, ‘FlexGen’ focuses on enabling high-throughput generative reasoning for large language models using a single GPU. The system employs a flexible memory and compute resource management strategy that integrates GPU, CPU and disk memory hierarchies to optimise performance. Notably, FlexGen employs an adaptive policy selection mechanism to efficiently adapt to various hardware configurations and model requirements. When running on a single GPU, FlexGen achieves a more significant improvement in token generation rate compared to traditional systems.

In previous work, a number of scholars have optimised KV Cache by token dropping methods. Scissorhands\cite{liu2024scissorhands} and H2O (Heavy-Hitter Oracle) \cite{zhang2024h2o} are two state-of-the-art methods that significantly reduce the memory usage of the KV cache. The Scissorhands system is based on the ‘importance persistence assumption’ and achieves up to 5x memory usage reduction without the need to fine-tune the model, by tracking and retaining the key tokens that have a significant impact on future generation. The Scissorhands system is based on the ‘Importance Persistence Assumption’ and achieves up to a 5x reduction in memory usage without the need to fine-tune the model, and can be combined with 4-bit quantization to further compress the memory without compromising the quality of the model while reducing memory usage.The H2O method, on the other hand, discovers that a small fraction of important words (Heavy Hitters) contribute most of the value by looking at the attention scores. The method solves the KV cache elimination problem by using a dynamic retention strategy of recent and Heavy Hitters tokens, significantly improves inference efficiency and throughput, and validates its accuracy and efficiency on tasks such as OPT, LLAMA, and GPT-NeoX. $H_2O$ significantly reduces memory footprint and improves performance in a variety of cases, while combining with quantization methods works particularly well . Both methods effectively reduce memory bottlenecks in LLM deployments while maintaining model accuracy.

KV cache compression method can be divided into KV cache quantization and KV cache compression.

In terms of KV cache quantization, recent studies include KVQuant\cite{hooper2024kvquant}, KIVI\cite{liu2024kivi}, QAQ\cite{dong2024qaq}, SKVQ\cite{duanmu2024skvq}, WKVQuant\cite{yue2024wkvquant}, and L2 Norm-Based Strategy\cite{devoto2024simple}

Experiments by KVQuant\cite{hooper2024kvquant} and KIVI\cite{liu2024kivi} have proved that key matrices often have distinct outlier channels. These outlier channels significantly impact other channels when quantizing the key matrices. To mitigate these impacts, KVQuant and KIVI attempt to quantize the matrices along the channel direction, which is called per-cahnnel. Besides the outlier problem, the rotary position embedding(RoPE) will cause mixing pairs of channels by different amounts for different position in the sequence. To handle the RoPE challenge, KVQuant develop a fused kernel to dequantize the pre-RoPE quantization and efficiently aplly the posisional embedding after that. For value matrices, there exists both outlier channels and outlier tokens, but they are much less than the outlier key channels. KIVI also finds that when the value matrix is per-channel quantized, the accuracy significantly decreases regardless of the quantization precision, and the most accurate quantization approach is to quantize key matrices per-channel and value matrices per-token when quantizing by a low numerical precision such as INT2. Besides that, KIVI maintains a full precision KV cache sliding window for the near tokens. Only the tokens are outside the sliding window are quantized to maintain the accuracy.

QAQ\cite{dong2024qaq} finds through partial derivative derivation and experimentation that the key cache is more sensitive to quantization, leading to more severe performance degradation when quantized. Consequently, QAQ proposed using different quantization methods and dynamic precisions for key cache and value cache. Additionally, QAQ finds exceptions to the 'Importance Persistence Assumption.' When performing quantization based on this hypothesis without additional treatment for exceptional cases, It could significantly impact the model's performance. QAQ proposes the attention window as the additional treatment for the exceptional cases, which maintains a slide window of size $n$ for attention, and only quantizes it to lower bits if the attention scores within these $n$ windows remain consistently low.

SKVQ proposes a solution to the low-bitwidth KV cache quantization problem, which mainly solves the problem of low-bitwidth KV cache quantization, i.e., how to improve the quantization performance and reduce the quantization error under the premise of guaranteeing the computational efficiency. Firstly, a channel reordering strategy is introduced to group channels with similar data distribution and then quantise them together, thus reducing the quantization error. Then, a clipping dynamic quantization strategy is proposed, which introduces a clipping factor in the quantization process, which can effectively reduce the influence of outliers in the channel, and realise the efficient improvement of the quantization performance through offline calibration. Next, this paper proposes a sliding-window quantization strategy, which utilises the KV cache obtained from full-precision computation in the pre-population stage, and retains a certain number of token pairs at the end to be processed with high precision, which can achieve significant performance improvement in long sequence tasks while adding little extra overhead.

WKVQuant can effectively reduce the memory footprint of pre-training large-scale language models while maintaining high model accuracy. This helps improve the performance of language models deployed in resource-constrained environments such as mobile devices. WKVQuant employs a Past-Only quantization (POQ) approach to improve the accuracy of attention computation by quantising only the past KV caches at the decoding stage; it then employs a 2D quantization strategy that combines static channel smoothing with dynamic marker level pinpointing, which further reduces the error; finally, Cross-block Reconstruction Regularization (CRR) is introduced to optimise the parameters by calculating the difference between subsequent layers to avoid the bias problem caused by local reconstruction loss, and using Mean Absolute Error (MAE). Absolute Error (MAE) instead of MSE to reduce the effect of outliers.

Alessio Devoto et al. proposed a KV cache compression method based on the L2 paradigm number of key embeddings. By observing the attention distributions and L2 norms of key embeddings in different heads and layers, the research team found that key embeddings with low L2 norms are usually associated with higher attention scores. Therefore, the authors concluded that the size of the KV cache can be reduced by compressing key embeddings with higher L2 paradigms and reducing the impact on the output.

In terms of KV cache compression, recent studies include MLKV\cite{zuhri2024mlkv}, CLA\cite{brandon2024reducing}, KEYFORMER\cite{adnan2024keyformer}, GEAR\cite{kang2024gear}, and MiniCache\cite{liu2024minicache}.

MLKV can share KV heads between heads in the same layer and other layers. m is the number of layers that have their respective KV heads, and the size of the KV cache is 2bsmgdk when MLKV is used. In addition, this paper proposes a sliding-window quantization (SWQ) strategy, which applies a pre-population stage to the computed KV cache with pre-population stage with full precision and then retaining a certain number of token pairs at the end for high-precision processing, which can obtain significant performance improvement in long sequence tasks while adding little extra overhead, and mainly solves the quantization problem of KV cache with low bit-width.

Reducing Transformer Key-Value Cache Size with Cross-Layer Attention (CLA) mainly solves the memory consumption problem of large-scale pre-trained models in natural language processing tasks. CLA makes it possible to perform only one key/value projection between the same set of neighbouring layers by sharing the key/value header, which greatly reduces the memory occupation and computation. In addition, CLA can be used in conjunction with other attention mechanisms to improve model effectiveness.

M Adnan et al. proposed the KEYFORMER approach, which aims to address the problem of increasing KV cache size in large language models.KEYFORMER utilises a sparsity approach to reduce the KV cache size. By exploiting the inherent sparsity in the attention mechanism, shorter subsequences can be selected to reduce the size of the KV cache and improve reasoning efficiency. This approach is important for tasks that require processing long texts, saving computational resources and accelerating the text generation process.

GEAR saves memory resources by compressing the KV cache matrix. Traditional approaches can only apply one compression technique alone, such as quantization, singular value decomposition or sparsification, but these methods cannot consider different information types simultaneously. In contrast, GEAR uses an integrated compression strategy that combines three different compression techniques to handle different types of information more efficiently and thus achieve better compression of the KV cache matrix to save memory resources.GEAR first uses a filter to extract the maximum and minimum values in the input tensor X and stores them in the sparse matrix S. The matrix is then sparsified by the filter. Then, the extracted matrix is uniformly quantised to obtain the quantised backbone D. Finally, the singular value decomposition algorithm is used to compute the remaining residual matrix R and its first r singular vectors are used to construct the low-rank matrix L. This integrated compression strategy allows GEAR to capture and recover information more efficiently, resulting in higher compression ratios and smaller approximation errors.

Liu et al. proposes a cross-layer KV cache compression method,which is called MiniCache. This idea is inspired by the prior studies, which revealed the ineffectiveness in the middle-to-deep layer in LLMs. It is a layer-wise merging method of KV cache. They had two observation: One is that through KV cache shares a high similarity between adjacent layers but they may have different magnitudes. Scaling is needed before merging the similar KV cache. and the other is that not all tokens are equally important to merge, a few distinct part require retention. Minicache merges the adjacent layers in the middle-to-deep layer after scaling them and retain a few tokens which are sensitive distinct. When inference, minicache first rescales the merged KV cache with the corresponding magnitude along the token dimension and places the sensitive tokens according to their token indices.

In the area of KV Cache management and optimisation techniques, recent research contains Cached Transformers\cite{zhang2024cached}, Layer-Condensed KV Cache \cite{wu2024layer}, and InfiniGen \cite{lee2024infinige}

Cached Transformers solves the problems of high computational complexity and difficulty in capturing long-term dependencies faced by traditional self-attentive mechanisms when dealing with long sequence data.The Cached Transformers approach combines current input samples with historical samples and achieves efficient long sequence modelling by dynamically recording historical samples and using them as a cache. In addition, the method employs a gated recurrent unit (GRU) update rule to dynamically capture dependencies on different time scales. The approach also provides a more flexible paradigm that can be used to develop applications such as cross-task memory modules.

The Layer-Condensed KV Cache method solves two main problems: reducing memory consumption and computation and maintaining information transfer between different layers. The method discards its own attention term in the self-attention of each token, which is equivalent to masking the main diagonal of the attention matrix. As a result, the first token does not have any self-attention term to refer to and simply uses the zero vector as a virtual key-value pair in its attention computation. Despite the fact that each token has no self-attention term, the information from all tokens can still be taken into account by means of bottom-up computation through the information transfer of residual connections. By pairing queries from all layers with key-value pairs using only the top layer, the computation of key-value pairs from all layers except the top layer is avoided, thus reducing memory consumption and computation. To solve the circular dependency problem in self-attention, the method uses a mask to handle the main diagonal so that the first token does not depend on itself but still maintains the effective delivery of information. Meanwhile, to address the fact that applying key-value pairs at the same level may undermine the previously observed tendency of focusing on syntactic information at lower levels and semantic information at higher levels, the paper proposes a strategy of retaining some of the ‘warmer layers’ in order to maintain the information transfer between different levels.

W Lee et al. proposed the InfiniGen method, which mainly addresses the data transfer overhead of loading and computing key value caches in large-scale pre-trained models.InfiniGen introduces a predictive cache module which is capable of modifying weight matrices online in order to generate skewed query and key matrices, which improves prediction accuracy and efficiency. In addition, InfiniGen employs a CPU memory pool management strategy to automatically select and replace the least important KV entries when a user-defined memory limit is reached.

\subsection{Quantization-Aware Training (QAT) and Post-Training Quantization (PTQ)}

Quantization is a vital technique in the field of deep learning aimed at reducing the model size and improving inference speed without significantly compromising accuracy. Two primary methods for quantization are Quantization-Aware Training (QAT) and Post-Training Quantization (PTQ).

Quantization-Aware Training (QAT) involves incorporating quantization into the training process itself. During QAT, the model is trained while simulating the effects of quantization, allowing the model to learn and adapt to the quantized weights and activations. This method typically yields higher accuracy compared to PTQ, as the model adjusts its parameters to minimize the loss function considering the quantization effects. QAT is particularly beneficial when the target hardware has stringent performance constraints, and maintaining high accuracy is crucial.

Post-Training Quantization (PTQ), on the other hand, is applied after the model has been fully trained. It involves converting the weights and activations of the pre-trained model from floating-point precision (usually 32-bit) to lower precision formats like 8-bit integers. PTQ is easier to implement since it doesn’t require retraining the model and can be applied to a wide range of pre-trained models. However, the accuracy might be slightly lower than QAT, especially for models that are sensitive to quantization errors.

Both QAT and PTQ offer significant benefits in terms of reduced model size and increased inference speed, making them essential techniques for deploying deep learning models on resource-constrained devices like mobile phones and embedded systems. The choice between QAT and PTQ depends on the specific requirements of the application, such as the importance of model accuracy, available computational resources, and deployment constraints.

\subsubsection{PTQ}
Post-Training Quantization (PTQ) is a technique for compressing deep neural networks by reducing the numerical precision of weights and activations after the model has been trained. Here is a survey of related works on PTQ:
PTQ aims to quantize a pre-trained full-precision model without retraining or access to the original training data. Early works like ~\cite{liu2021post}  proposed multipoint quantization, which approximates full-precision weights as a linear combination of multiple low-bit vectors, allowing flexible trade-offs between accuracy and model size on a per-channel basis.
More recent works have explored PTQ for vision transformers like ~\cite{liu2021post,liu2023pd}. PTQ-ViT~\cite{liu2021post}  proposes similarity-aware quantization for linear layers and ranking-aware quantization for self-attention layers in vision transformers. It also uses mixed-precision based on the nuclear norm of attention maps and output features. PTQ4ViT~\cite{liu2023pd}  introduces techniques like twin uniform quantizers and Hessian-guided metrics for quantizing vision transformer activations.
Several works have focused on improving PTQ accuracy, especially for low bit-widths (< 8 bits)~\cite{hubara2021accurate}.  proposes a three-stage pipeline: 1) minimizing per-layer quantization error on a small calibration set, 2) allocating bit-widths optimally via integer programming, and 3) tuning global model statistics. This achieves state-of-the-art results like 4-bit quantized ResNet50 with $< 1\%$ accuracy drop.
Other notable PTQ works include  which surveys techniques for transformer compression including quantization methods tailored for vision transformers. Outlier suppression  is proposed to handle outliers when quantizing transformers.
In summary, PTQ has seen significant research interest, with works proposing novel quantization schemes for vision transformers, improving accuracy for low bit-widths, and optimally allocating bit-widths across layers.

~\cite{Banner2018PostT4} propose three novel methods for improving quantization. First, ACIQ clips the activation range to an optimal value determined by minimizing the mean squared error(MSE), thereby reducing quantization error in regions with the highest information density. Second, per-channel bit allocation determines the optimal bit-width for each channel to minimize overall MSE. Third, Bias Correction compensates for the inherent bias and variance changes in weight after quantization.

DFQ ~\cite{Nagel2019DataFreeQT} propose a data-free method for 4-bit quantization. This approach equalizes the weight ranges across the network by leveraging the scale-equivariance property of activation functions, synchronizing the weights of multiple layers, and integrating high biases into the subsequent layer. Additionally, the method includes bias correction during quantization.

LBQ ~\cite{Choukroun2019LowbitQO} minimize the MSE by optimizing each network layer individually and utilizing multiple quantization tensors to address key layers with high MSE. Additionally, they propose a method for post-quantization adjustment of scaling factors, which optimizes a small number of parameters for better approximation of the original model.

PWLQ ~\cite{Fang2020PosttrainingPL} introduces a piece-wise linear quantization scheme that enhances precision through the strategic division of the quantization range into two non-overlapping regions with equal quantization levels.To find the optimal breakpoints, the authors minimize the quantization error. Additionally, bias correction is applied to further refine accuracy.

SPARQ ~\cite{Shomron2021PostTrainingSQ} dynamically selects the most significant bits from 8-bit activations, bypassing leading zeros, and quantizes pairs of activations to 4 bits, using the full 8-bit budget if one activation is zero. Implemented on a systolic array and Tensor Core DP unit, SPARQ demonstrates low area overhead while maximizing computational efficiency.

EasyQuant ~\cite{Tang2024EasyQuantAE} introduces a data-free weight-only quantization algorithm, which optimizes the quantization range using a gradient-based method while strategically leaving less than 1\% of outliers unchanged to minimize quantization error and reduce reconstruction error. The algorithm's ability to be implemented in parallel makes it highly efficient for LLMs over 100B. Remarkably, EasyQuant outperforms traditional data-dependent methods, delivering over 10 times faster performance.

BRECQ ~\cite{Li2021BRECQPT} advances the frontiers of neural network quantization by enabling INT2 precision for the first time. By reconstructing the fundamental building blocks of neural networks one at a time and optimizing cross-layer dependencies through second-order error analysis, BRECQ achieves a balance between precision and generalization. The integration of mixed precision techniques and genetic algorithm further enhances the efficacy of quantization. BRECQ can produce 4-bit ResNet and MobileNetV2 models that rival the performance of QAT, with a 240x faster production of quantized models.

PTQD ~\cite{He2023PTQDAP} introduces a novel quantization approach for diffusion models by disentangling quantization noise into correlated and residual uncorrelated components. The linearly correlated part is mitigated through the estimation of the correlation coefficient, while the latter is treated as a Gaussian distribution, addressed by subtracting the bias and modifying the variance schedule. Additionally, a step-aware mixed-precision scheme is employed to optimize bitwidth allocation, effectively maintaining a high SNR while significantly reducing computational complexity. This approach not only reduces the FID gap to just 0.06 in 250 generation steps under a W4A8 bitwidth setting but also compresses the model size by 6.83× and cuts bit operations by 19.96×, demonstrating its superiority over existing models. 

ZeroQ ~\cite{Cai2020ZeroQAN} introduces a novel quantization approach that generates a distilled dataset aligned with network batch normalization statistics, enabling effective quantization without requiring training data. This method supports both uniform and mixed-precision quantization, automatically determining the optimal bit settings using a Pareto frontier approach. 

\definecolor{MidnightBlue}{HTML}{006795}
\definecolor{RedOrange}{HTML}{F26035}
\definecolor{Plum}{HTML}{92268F}
\definecolor{Dandelion}{HTML}{00CD00}
\definecolor{Green}{HTML}{2E8B57}
\definecolor{Orange}{HTML}{FF8C00}
\definecolor{light-sky-blue}{RGB}{154,180,205}
\definecolor{klein-blue}{RGB}{0,47,167}

\begin{figure*}[htbp]
\centering
\begin{tikzpicture}[
    yearblock1/.style={rectangle, draw=none, fill=MidnightBlue!20,minimum width=3cm, minimum height=1cm, align=center, font=\large\bfseries},
    yearblock2/.style={rectangle, draw=none, fill=RedOrange!20,minimum width=3cm, minimum height=1cm, align=center, font=\large\bfseries},
    yearblock3/.style={rectangle, draw=none, fill=Plum!20,minimum width=3cm, minimum height=1cm, align=center, font=\large\bfseries},
    yearblock4/.style={rectangle, draw=none, fill=Dandelion!20,minimum width=3cm, minimum height=1cm, align=center, font=\large\bfseries},
    yearblock5/.style={rectangle, draw=none, fill=Green!20,minimum width=3cm, minimum height=1cm, align=center, font=\large\bfseries},
    yearblock6/.style={rectangle, draw=none,minimum width=3cm, minimum height=1cm, align=center, font=\large\bfseries, shading=axis,left color=klein-blue!20,right color=klein-blue!10 },
    paper/.style={text width=3cm, align=left, anchor=west, font=\small},
]

\node[yearblock1] (2019) at (0,0) {\textcolor{MidnightBlue}{2019}};

\node[yearblock2] (2020) at (3,0) {\textcolor{RedOrange}{2020}};

\node[yearblock3] (2021) at (6,0) {\textcolor{Plum}{2021}};

\node[yearblock4] (2022) at (9,0) {\textcolor{Dandelion}{2022}};

\node[yearblock5] (2023) at (12,0) {\textcolor{Green}{2023}};

\node[yearblock6] (2024) at (15,0) {\textcolor{klein-blue}{2024}};

\fill[klein-blue!10] (16.5,0.5) -- (17,0) -- (16.5,-0.5) -- cycle; 

\node[paper] (lbq) at (-0.75,-2) {LBQ ~\cite{Choukroun2019LowbitQO}};
\node[paper] (aciq) at (-0.75,-1.5) {ACIQ ~\cite{Banner2018PostT4}};
\node[paper] (dfq) at (-0.75,-1) {DFQ ~\cite{Nagel2019DataFreeQT}};

\node[paper] (pwlq) at (2.25,-1) {PWLQ ~\cite{Fang2020PosttrainingPL}};
\node[paper] (adaround) at (2.25,-1.5) {AdaRound ~\cite{Nagel2020UpOD}};
\node[paper] (zeroq) at (2.25,-2) {ZeroQ ~\cite{Cai2020ZeroQAN}};

\node[paper] (adaquant) at (5.25,-1) {AdaQuant ~\cite{hubara2021accurate}};
\node[paper] (brecq) at (5.25,-1.5) {BRECQ ~\cite{Li2021BRECQPT}};
\node[paper] (sparq) at (5.25,-2) {SparQ ~\cite{Shomron2021PostTrainingSQ}};
\node[paper] (ptqvit) at (5.25,-2.5) {PTQ-ViT ~\cite{liu2021post}};

\node[paper] (zeroquant) at (8.25,-1) {ZeroQuant ~\cite{yao2022zeroquant}};
\node[paper] (lutgemm) at (8.25,-1.5) {LUT-GEMM ~\cite{park2022lut}} ;
\node[paper] (smoothquant) at (8.25,-2) {SmoothQuant ~\cite{xiao2023smoothquant}};
\node[paper] (gptq) at(8.25,-2.5) {GPTQ ~\cite{frantar2022gptq}};

\node[paper] (ptqd) at (11.25,-1) {PTQD ~\cite{He2023PTQDAP}};
\node[paper] (llmqat) at (11.25,-1.5) {\textcolor{red}{LLM-QAT } ~\cite{liu2023llm}};
\node[paper] (qlora) at (11.25,-2) {\textcolor{red}{QLORA } ~\cite{dettmers2024qlora}};
\node[paper] (olive) at (11.25,-2.5) {OliVe ~\cite{guo2023olive}};
\node[paper] (spqr) at (11.25,-3) {SpQR ~\cite{dettmers2023spqr}};
\node[paper] (awq) at (11.25,-3.5) {AWQ ~\cite{lin2024awq}};
\node[paper] (pbllm) at (11.25,-4) {\textcolor{red}{PB-LLM} ~\cite{shang2023pbllmpartiallybinarizedlarge}};
\node[paper] (ptq4vit) at (11.25,-4.5) {PTQ4ViT ~\cite{liu2023pd}};

\node[paper] (easyquant) at (14.25,-1) {EasyQuant ~\cite{Tang2024EasyQuantAE}};
\node[paper] (l4q) at (14.25,-1.5) {\textcolor{red}{L4Q }~\cite{jeon2024l4q} };
\node[paper] (edgeqat) at (14.25,-2) {\textcolor{red}{EdgeQAT} ~\cite{shen2024edgeqatentropydistributionguided}};
\node[paper] (quarot) at (14.25,-2.5) {\textcolor{red}{QuaRot} ~\cite{ashkboos2024quarot}};
\node[paper] (spinquant) at (14.25,-3) {\textcolor{red}{SpinQuant} ~\cite{liu2024spinquantllmquantizationlearned}};
\node[paper] (lrqat) at (14.25, -3.5) {\textcolor{red}{LR-QAT} ~\cite{bondarenko2024lowrankquantizationawaretrainingllms}};
\node[paper] (efficientqat) at (14.25,-4) {\textcolor{red}{EfficientQAT} ~\cite{chen2024efficientqatefficientquantizationawaretraining}};

\end{tikzpicture}
\caption{Timeline of QAT and PTQ. The \textcolor{red}{red} highlighted methods represent they belonging to QAT-related methods, and others are PTQ-based methods.}
\end{figure*}




\subsubsection{QAT}
quantization Aware Training (QAT) is a model compression and acceleration technique that takes quantization into account during the training process. Different from the traditional post-training Quantization, QAT simulates the quantization operation in the training phase, so that the model can adapt to the information loss caused by quantization in the optimisation process. Specifically, QAT inserts ‘pseudo-quantization nodes’ into the model, i.e., simulated quantization and anti-quantization of weights and activation values, so that the model is aware of the quantization error during training. This allows the model to be aware of the quantization error during training, so that the model can maintain high performance even when low-precision integer arithmetic is used in the actual deployment. Compared with post-quantization alone, QAT can better balance model size, inference speed and accuracy, which is especially important for large-scale language models deployed on resource-constrained devices.

quantization Aware Training was first introduced by Google researcher Benoit Jacob et al. at ICLR 2018 in a paper ‘Quantization and Training of Neural Networks for Efficient Integer-Arithmetic-Only Inference’ \cite{jacob2017quantizationtrainingneuralnetworks}, which first introduced the concept of quantization Aware Training. They introduced a method to simulate quantization operations during the training process in neural networks, so that the model can adapt to the precision loss caused by quantization, and thus use low-precision integer operations without significant performance degradation when deployed in practice.

Later, the QAT method was gradually widely used in various fields such as quantization optimisation for neural network training and inference \cite{esser2020learnedstepsizequantization}, and quantization of graph neural networks \cite{gao2022efficientgraphneuralnetwork}.

\textbf{LLM-QAT.} LLM-QAT\cite{liu2023llm}is the first application of QAT techniques to large models. The authors propose a novel data-free distillation method to apply quantitative awareness training (QAT) to large language models (LLMs). The authors analyse the difficulties of quantitative training for LLM: firstly, it is very important to choose the appropriate fine-tuning data, which may damage the model performance if the fine-tuning data does not match the original pre-training data distribution or is too narrow; secondly, due to the complexity and huge scale of LLM pre-training, it is difficult to replicate the original training setup exactly; furthermore, the weight and activation distributions of LLM are significantly different from those of small-scale models, and thus are applicable to the LLM. To address these difficulties, the paper proposes a data-free distillation method, which uses the original pre-training model to generate the next token data as QAT training data, and adopts a hybrid sampling strategy, where the first 3-5 tokens are sampled using top prediction, and the subsequent tokens are sampled using random sampling based on output probability, and the first 3-5 tokens are sampled using top prediction, and the subsequent tokens are sampled using random sampling based on output probability, and the next tokens are sampled using random sampling based on output probability. The first 3-5 tokens are sampled using top prediction, and the subsequent tokens are sampled using random sampling based on output probability, and synthetic data are generated using GAN for QAT training. In the quantization method, the paper chooses the symmetric MinMax linear quantization instead of cropping quantization, because there are a large number of outliers in the LLM, and cropping these values will seriously damage the model performance. In addition, the quantization technique is applied to KV cache, a key component of LLM, for the first time, which further compresses the model size.

In the recent study, we have categorised and summarised the thesis research directions.

\textbf{Data-free QAT.}This category of methods aims to perform QAT training by generating data or other techniques without relying on the original training data. LLM-QAT\cite{liu2023llm} also belongs to this class of QAT quantization methods. Meanwhile, EdgeQAT \cite{shen2024edgeqatentropydistributionguided} is also a Data-free QAT quantization method. It is an innovative approach to optimise the quantization process by introducing an adaptive quantization strategy. Specifically, EdgeQAT will adopt different quantization parameters according to the characteristics of different layers. This adaptive quantization approach can better balance the model performance and model size, thus enabling the deployment of low-bit quantization models on edge devices while maintaining high inference performance. Unlike the traditional one-size-fits-all quantization approach, EdgeQAT's adaptive quantization strategy can make full use of the characteristics of different layers. Some critical layers can retain higher bits, while non-critical layers can be quantised to lower bits. This differentiated quantization minimises the performance loss caused by quantization, so that the low-bit quantization model can still maintain high accuracy and inference speed on edge devices. This Data-free type of QAT method does not need to rely on the original training data, but achieves the efficient compression of the model through adaptive quantization and other techniques, which provides new possibilities for the deployment of LLM models on resource-constrained edge devices. Compared with the traditional QAT methods, data free methods such as EdgeQAT are more flexible and general, and can be widely applied to different types of LLM models.

\textbf{Matrix transform-based QAT.}This type of method optimises the quantization process and improves the performance by applying specific matrix transformations (e.g. rotations) to the model weights or activations. This approach should belong to the new class of quantization techniques that make other modifications to QAT. Representative works include QuaRot \cite{ashkboos2024quarot} and SpinQuant \cite{liu2024spinquantllmquantizationlearned}, which are two matrix-transformation-based LLM quantization methods that optimize the quantization process by applying specific transformations to the model weights and activations to mitigate the performance degradation caused by quantization. SpinQuant proposes a quantization framework based on the spin system, which encodes the model parameters into the states of the spin system to achieve more efficient quantization, and this method can significantly improve the performance of the quantised model.QuaRot optimises the quantization process by applying orthogonal transformations to the model weights and activations. The orthogonal transformation can maintain the expressive power of the model and avoid the performance degradation caused by quantization. Specifically, QuaRot performs orthogonal transformations on weights and activations before quantization. The orthogonal transformation ensures that the quantised model still maintains good expressive ability, thus improving the performance after quantization. Compared with the traditional quantization methods, these two matrix transformation-based methods can better adapt to the quantization process by applying specific transformations to the weights and activations, thus mitigating the performance degradation caused by quantization. This is important for deploying high-performance LLM models on resource-constrained edge devices.

\begin{table*}[ht]
    \centering
    \renewcommand{\arraystretch}{1.5} 
    \setlength{\tabcolsep}{10pt} 
    \begin{tabularx}{\textwidth}{p{0.35\textwidth} 
  p{0.65\textwidth}}
        \hline
        \textbf{QAT Category} & \textbf{Details} \\ \hline
        
        Data free QAT method & 
        \begin{itemize}
            \item LLM-QAT
            \item EdgeQAT
        \end{itemize} \\ \hline
        
        Matrix transform-based QAT & 
        \begin{itemize}
            \item QuaRot
            \item SpinQuant
        \end{itemize} \\ \hline
        
        Lightweight and efficient QAT approach & 
        \begin{itemize}
            \item LR-QAT
            \item EfficientQAT
        \end{itemize} \\ \hline
        
        Extreme low-bit quantisation methods & 
        \begin{itemize}
            \item PB-LLM
        \end{itemize} \\ \hline

    \end{tabularx}
    \caption{Summary of QAT.}
    \label{tab:qat_methods}
\end{table*}


\textbf{Lightweight and Efficient QAT.} Such methods aim to improve the efficiency and practicality of QAT and reduce the consumption of training resources. Representative works include LR-QAT \cite{bondarenko2024lowrankquantizationawaretrainingllms}, EfficientQAT \cite{chen2024efficientqatefficientquantizationawaretraining}, and L4Q \cite{jeon2024l4q}. LR-QAT introduces a low-rank adaptation technique, which dramatically reduces the memory and computational overheads of QAT by performing low-rank decomposition of model weights. The traditional QAT method needs to save the quantization parameters for each weight parameter individually, which brings huge memory overhead. In contrast, LR-QAT compresses the weight parameters into low-rank representations through low-rank decomposition, which significantly reduces the memory and computational resources required for QAT. At the same time, it also adopts the block training strategy, which further improves the efficiency of QAT. These innovations make LR-QAT a lightweight and efficient QAT method. EfficientQAT, on the other hand, introduces an adaptive quantization strategy to adopt different quantization parameters according to the characteristics of different layers. As the characteristics of different layers are quite different, it is difficult to balance the performance of each layer by using a uniform quantization parameter. EfficientQAT addresses this problem by designing an adaptive quantization parameter for each layer, thus better balancing the performance of each layer. In addition, EfficientQAT also makes use of the knowledge distillation technique to further improve the efficiency and practicability of QAT. L4Q proposes a parameter-efficient quantization-aware fine-tuning method, which can significantly reduce the model size while maintaining the model performance, effectively balancing the model compression and performance preservation. "How to Parameterize Asymmetric Quantization Ranges for Quantization-Aware Training", on the other hand, investigates how to parameterize asymmetric quantization ranges to improve the effectiveness of quantization-aware training, so as to better balance the performance of different layers and enhance the overall model performance. It improves the quantization effect by introducing learnable asymmetric quantization parameters to better adapt to the differences in the characteristics of different layers. Specifically, the method introduces two learnable scaling factors $\alpha$ and $\beta$ in the quantization process, which are used to scale the quantization range of positive and negative weights respectively. This can adaptively adjust the quantization ranges of different layers to better capture the asymmetry of the weight distribution. Meanwhile, the authors also propose a gradient-based optimization method, which can optimize the network parameters and quantization parameters simultaneously during the training process to further improve the quantization performance. This parametric asymmetric quantization method belongs to the lightweight and efficient QAT category, which can significantly reduce the model size and computational overhead while maintaining the model performance. Compared with traditional symmetric quantization, this method can better adapt to the differences in the characteristics of different layers, thus improving the quantization effect.

\textbf{Extreme low-bit quantization.} This class of methods explores quantising models to extremely low bit counts (e.g., 2-4bit) while maintaining performance as much as possible. Representative works include PB-LLM \cite{shang2023pbllmpartiallybinarizedlarge}.PB-LLM is an extreme low-bit quantization method for LLM models, which achieves the goal of maintaining good performance at very low bits through two key innovations.PB-LLM has two innovations. First, the first innovation is the partial preservation of significant weights. Traditional quantization methods quantise all weight parameters to the same low bit-width, which results in a significant degradation of model performance. In contrast, PB-LLM identifies the critical weight parameters in the model and retains them in the original high bit-width, quantising only the non-critical weights. This strategy of partially retaining the critical weights allows the model to maintain a good inference performance at very low bits. The second innovation of the article is the optimisation of the quantization parameters.PB-LLM designs adaptive quantization parameters for each layer with respect to the differences in the characteristics of different layers. This strategy can better balance the performance of each layer and avoid the problems that may be caused by using uniform quantization parameters. In addition, PB-LLM further optimises the quantization parameters by using knowledge distillation and other techniques to improve the quantization effect. Compared with the traditional 8-bit or 4-bit quantization method, PB-LLM can further compress the model size to a lower bit width while maintaining a higher inference performance. This extreme low-bit quantization method enables excellent models to be deployed in common hardware conditions and facilitates model generalisation.

These approaches demonstrate the properties of QAT technology: the ability to be deployed in low-end hardware and to be quantised during training, reducing the loss of model accuracy.

\subsubsection{PTQ}
The core idea of PTQ is to quantize model weights and activations after training, thus reducing the computational and memory footprint without further modifying the model. The seminal work by Jacob et al. (2018), titled "Quantization and Training of Neural Networks for Efficient Integer-Arithmetic-Only Inference," lays the foundation for modern PTQ techniques. The authors introduced linear quantization methods that map floating-point values to integers, enabling convolutional neural networks to run efficiently on hardware that only supports integer operations. This work has been widely adopted and has significantly influenced subsequent developments in PTQ.

\textbf{ Practical Applications and Challenges of PTQ.} Migacz (2017) in "Integer Quantization for Deep Learning Inference: Principles and Empirical Evaluation" provided a practical evaluation of PTQ in real-world scenarios. The paper analyzed the impact of quantization precision on model inference performance and offered insights into how to choose appropriate quantization strategies for different applications. This research serves as a valuable guide for engineers deploying deep learning models on mobile devices and in edge computing environments.
However, a major challenge of PTQ is the loss of accuracy introduced by quantization. While PTQ does not require retraining, direct quantization can lead to significant performance degradation, particularly in tasks that demand high precision. This accuracy loss is often pronounced in certain layers or parts of the model, necessitating more refined quantization strategies.

\textbf{Advanced PTQ Optimization Strategies.}To address the accuracy loss associated with traditional PTQ methods, researchers have proposed various optimization strategies. Dong et al. (2019) introduced HAWQ: Hessian AWare Quantization of Neural Networks with Mixed-Precision, a novel approach that uses Hessian matrix information to guide mixed-precision quantization. By computing the Hessian matrix, this method identifies layers that are sensitive to quantization and assigns higher precision to them, thus maintaining overall model accuracy while reducing computational costs.
Cai et al. (2020), in "Towards Accurate Post-Training Network Quantization via Bit-Split and Stitching," proposed another method to enhance quantization accuracy. They introduced a technique that splits weights into smaller segments, quantizes them separately, and then stitches them back together, effectively reducing quantization error. This approach performs well in applications requiring high-precision inference.
Additionally, Choi et al. (2018) proposed PACT: Parameterized Clipping Activation for Quantized Neural Networks, a technique that optimizes the quantization process by adjusting the range of activation values. The PACT method uses parameterized clipping to better align activation values with quantization, thereby reducing accuracy loss during inference.

\textbf{Q-DiT.}Building upon the foundational work in PTQ, recent advancements have focused on addressing the unique challenges posed by complex model architectures like Diffusion Transformers. A notable contribution in this area is the Q-DiT framework, which introduces several innovative techniques to enhance the accuracy and efficiency of post-training quantization for these models.
Q-DiT leverages a combination of fine-grained group quantization and dynamic activation quantization to tackle the high variance and structured outliers in activations that are characteristic of Diffusion Transformers. These techniques ensure that the quantization process remains effective across the model's various layers, even under aggressive quantization settings like W4A8. Additionally, Q-DiT employs an evolutionary search algorithm to optimize quantization granularity, balancing computational efficiency with performance.
Experimental results on the ImageNet dataset validate the effectiveness of Q-DiT, showing that it can achieve near-lossless compression, setting a new standard for PTQ in high-performance generative models. These advancements underscore the importance of tailored PTQ strategies for different model architectures and pave the way for further innovations in the field.

\textbf{AdaLog.}Further extending the boundaries of PTQ, the AdaLog framework addresses the specific challenges of Vision Transformer (ViT) architectures, which are increasingly popular but computationally expensive. Traditional PTQ methods often struggle with the power-law distributions of post-Softmax and post-GELU activations in ViTs, leading to significant accuracy degradation, especially at low-bit quantization levels.
AdaLog introduces an Adaptive Logarithm Quantizer, which optimizes the logarithmic base used in the quantization process rather than relying on a fixed base. This adaptability allows AdaLog to better handle the unique activation distributions of ViTs, maintaining higher accuracy even under 3-bit and 4-bit quantization settings. The framework also incorporates bias reparameterization, enabling the effective quantization of both post-Softmax and post-GELU activations, and making the process more hardware-friendly.
To enhance the efficiency of the quantization process, AdaLog employs a Fast Progressive Combining Search (FPCS) strategy. This strategy efficiently determines the optimal hyperparameters for quantization, balancing precision with computational cost. Experimental results demonstrate that AdaLog significantly outperforms existing PTQ methods in various ViT architectures, providing a robust solution for deploying these complex models in resource-constrained environments.

\textbf{Future Research Directions.}Despite the success of PTQ in real-world applications, further reducing accuracy loss remains a key area of research. Future work may focus on:

\begin{itemize}
    \item Adaptive Quantization Strategies: Developing methods that dynamically adjust quantization precision based on different parts of the model to minimize quantization error.
    \item Mixed-Precision Quantization: Combining low-precision and high-precision computation to achieve the optimal balance between performance and efficiency.
    \item Combining PTQ with Quantization-Aware Training (QAT): Exploring methods to further optimize models with minimal retraining, improving both compression efficiency and accuracy.
\end{itemize}

Post-Training Quantization is an effective model compression technique, particularly suitable for deploying deep learning models in resource-constrained environments. While PTQ offers significant advantages in reducing computational and storage demands, the challenge of accuracy loss persists. The Q-DiT and AdaLog frameworks represent significant advances in addressing these challenges, particularly for Diffusion Transformers and Vision Transformers. Ongoing research into optimizing quantization strategies promises to enhance the efficacy of PTQ in a broader range of applications.





\section{Conclusion}
\label{SEC:Conclusion}
he quantization of large-scale neural network models emerges as a critical strategy in addressing the computational and energy demands associated with the growth of model sizes. Our comprehensive overview highlights the advancements in quantization techniques that enable significant reductions in model size and computational overhead while maintaining high levels of accuracy. By analyzing various algorithms and approaches, we observe that innovative methods such as LLM-QAT and SmoothQuant effectively balance performance and efficiency, making the deployment of large-scale models more feasible in resource-constrained environments. The integration of quantization-aware training and sophisticated post-training quantization methods demonstrates the potential for neural networks to continue scaling in complexity without proportional increases in computational costs. Future work in this domain is essential to further optimize quantization strategies, ensuring that the benefits of large-scale models can be widely accessible and environmentally sustainable.



{
\bibliographystyle{IEEEtran}
\bibliography{references.bib}

\begin{thebibliography}{10}
\providecommand{\url}[1]{#1}
\csname url@samestyle\endcsname
\providecommand{\newblock}{\relax}
\providecommand{\bibinfo}[2]{#2}
\providecommand{\BIBentrySTDinterwordspacing}{\spaceskip=0pt\relax}
\providecommand{\BIBentryALTinterwordstretchfactor}{4}
\providecommand{\BIBentryALTinterwordspacing}{\spaceskip=\fontdimen2\font plus
\BIBentryALTinterwordstretchfactor\fontdimen3\font minus \fontdimen4\font\relax}
\providecommand{\BIBforeignlanguage}[2]{{%
\expandafter\ifx\csname l@#1\endcsname\relax
\typeout{** WARNING: IEEEtran.bst: No hyphenation pattern has been}%
\typeout{** loaded for the language `#1'. Using the pattern for}%
\typeout{** the default language instead.}%
\else
\language=\csname l@#1\endcsname
\fi
#2}}
\providecommand{\BIBdecl}{\relax}
\BIBdecl

\bibitem{liu2024spinquantllmquantizationlearned}
\BIBentryALTinterwordspacing
Z.~Liu, C.~Zhao, I.~Fedorov, B.~Soran, D.~Choudhary, R.~Krishnamoorthi, V.~Chandra, Y.~Tian, and T.~Blankevoort, ``Spinquant: Llm quantization with learned rotations,'' 2024. [Online]. Available: \url{https://arxiv.org/abs/2405.16406}
\BIBentrySTDinterwordspacing

\bibitem{xiao2023smoothquant}
G.~Xiao, J.~Lin, M.~Seznec, H.~Wu, J.~Demouth, and S.~Han, ``Smoothquant: Accurate and efficient post-training quantization for large language models,'' in \emph{International Conference on Machine Learning}.\hskip 1em plus 0.5em minus 0.4em\relax PMLR, 2023, pp. 38\,087--38\,099.

\bibitem{dettmers2023spqr}
T.~Dettmers, R.~Svirschevski, V.~Egiazarian, D.~Kuznedelev, E.~Frantar, S.~Ashkboos, A.~Borzunov, T.~Hoefler, and D.~Alistarh, ``Spqr: A sparse-quantized representation for near-lossless llm weight compression,'' \emph{arXiv preprint arXiv:2306.03078}, 2023.

\bibitem{guo2023olive}
C.~Guo, J.~Tang, W.~Hu, J.~Leng, C.~Zhang, F.~Yang, Y.~Liu, M.~Guo, and Y.~Zhu, ``Olive: Accelerating large language models via hardware-friendly outlier-victim pair quantization,'' in \emph{Proceedings of the 50th Annual International Symposium on Computer Architecture}, 2023, pp. 1--15.

\bibitem{Tang2024EasyQuantAE}
H.~Tang, Y.~Sun, D.~Wu, K.~Liu, J.~Zhu, and Z.~Kang, ``Easyquant: An efficient data-free quantization algorithm for llms,'' \emph{ArXiv}, vol. abs/2403.02775, 2024.

\bibitem{frantar2022gptq}
E.~Frantar, S.~Ashkboos, T.~Hoefler, and D.~Alistarh, ``Gptq: Accurate post-training quantization for generative pre-trained transformers,'' \emph{arXiv preprint arXiv:2210.17323}, 2022.

\bibitem{dettmers2024qlora}
T.~Dettmers, A.~Pagnoni, A.~Holtzman, and L.~Zettlemoyer, ``Qlora: Efficient finetuning of quantized llms,'' \emph{Advances in Neural Information Processing Systems}, vol.~36, 2024.

\bibitem{yao2022zeroquant}
Z.~Yao, R.~Yazdani~Aminabadi, M.~Zhang, X.~Wu, C.~Li, and Y.~He, ``Zeroquant: Efficient and affordable post-training quantization for large-scale transformers,'' \emph{Advances in Neural Information Processing Systems}, vol.~35, pp. 27\,168--27\,183, 2022.

\bibitem{jeon2024l4q}
H.~Jeon, Y.~Kim, and J.-j. Kim, ``L4q: Parameter efficient quantization-aware training on large language models via lora-wise lsq,'' \emph{arXiv preprint arXiv:2402.04902}, 2024.

\bibitem{park2022lut}
G.~Park, B.~Park, M.~Kim, S.~Lee, J.~Kim, B.~Kwon, S.~J. Kwon, B.~Kim, Y.~Lee, and D.~Lee, ``Lut-gemm: Quantized matrix multiplication based on luts for efficient inference in large-scale generative language models,'' \emph{arXiv preprint arXiv:2206.09557}, 2022.

\bibitem{lin2024awq}
J.~Lin, J.~Tang, H.~Tang, S.~Yang, W.-M. Chen, W.-C. Wang, G.~Xiao, X.~Dang, C.~Gan, and S.~Han, ``Awq: Activation-aware weight quantization for on-device llm compression and acceleration,'' \emph{Proceedings of Machine Learning and Systems}, vol.~6, pp. 87--100, 2024.

\bibitem{banner2018aciq}
R.~Banner, Y.~Nahshan, E.~Hoffer, and D.~Soudry, ``Aciq: Analytical clipping for integer quantization of neural networks,'' 2018.

\bibitem{Choukroun2019LowbitQO}
Y.~Choukroun, E.~Kravchik, and P.~Kisilev, ``Low-bit quantization of neural networks for efficient inference,'' \emph{2019 IEEE/CVF International Conference on Computer Vision Workshop (ICCVW)}, pp. 3009--3018, 2019.

\bibitem{Nagel2019DataFreeQT}
M.~Nagel, M.~van Baalen, T.~Blankevoort, and M.~Welling, ``Data-free quantization through weight equalization and bias correction,'' \emph{2019 IEEE/CVF International Conference on Computer Vision (ICCV)}, pp. 1325--1334, 2019.

\bibitem{fang2020post}
J.~Fang, A.~Shafiee, H.~Abdel-Aziz, D.~Thorsley, G.~Georgiadis, and J.~H. Hassoun, ``Post-training piecewise linear quantization for deep neural networks,'' in \emph{Computer Vision--ECCV 2020: 16th European Conference, Glasgow, UK, August 23--28, 2020, Proceedings, Part II 16}.\hskip 1em plus 0.5em minus 0.4em\relax Springer, 2020, pp. 69--86.

\bibitem{Shomron2021PostTrainingSQ}
G.~Shomron, F.~Gabbay, S.~Kurzum, and U.~C. Weiser, ``Post-training sparsity-aware quantization,'' \emph{ArXiv}, vol. abs/2105.11010, 2021.

\bibitem{Li2021BRECQPT}
Y.~Li, R.~Gong, X.~Tan, Y.~Yang, P.~Hu, Q.~Zhang, F.~Yu, W.~Wang, and S.~Gu, ``Brecq: Pushing the limit of post-training quantization by block reconstruction,'' \emph{ArXiv}, vol. abs/2102.05426, 2021.

\bibitem{He2023PTQDAP}
Y.~He, L.~Liu, J.~Liu, W.~Wu, H.~Zhou, and B.~Zhuang, ``Ptqd: Accurate post-training quantization for diffusion models,'' \emph{ArXiv}, vol. abs/2305.10657, 2023.

\bibitem{Cai2020ZeroQAN}
Y.~Cai, Z.~Yao, Z.~Dong, A.~Gholami, M.~W. Mahoney, and K.~Keutzer, ``Zeroq: A novel zero shot quantization framework,'' \emph{2020 IEEE/CVF Conference on Computer Vision and Pattern Recognition (CVPR)}, pp. 13\,166--13\,175, 2020.

\bibitem{courbariaux2015binaryconnect}
M.~Courbariaux, Y.~Bengio, and J.-P. David, ``Binaryconnect: Training deep neural networks with binary weights during propagations,'' \emph{Advances in neural information processing systems}, vol.~28, 2015.

\bibitem{courbariaux2016binarized}
M.~Courbariaux, I.~Hubara, D.~Soudry, R.~El-Yaniv, and Y.~Bengio, ``Binarized neural networks: Training deep neural networks with weights and activations constrained to+ 1 or-1,'' \emph{arXiv preprint arXiv:1602.02830}, 2016.

\bibitem{rastegari2016xnor}
M.~Rastegari, V.~Ordonez, J.~Redmon, and A.~Farhadi, ``Xnor-net: Imagenet classification using binary convolutional neural networks,'' in \emph{European conference on computer vision}.\hskip 1em plus 0.5em minus 0.4em\relax Springer, 2016, pp. 525--542.

\bibitem{zhou2016dorefa}
S.~Zhou, Y.~Wu, Z.~Ni, X.~Zhou, H.~Wen, and Y.~Zou, ``Dorefa-net: Training low bitwidth convolutional neural networks with low bitwidth gradients,'' \emph{arXiv preprint arXiv:1606.06160}, 2016.

\bibitem{wu2018training}
S.~Wu, G.~Li, F.~Chen, and L.~Shi, ``Training and inference with integers in deep neural networks,'' \emph{arXiv preprint arXiv:1802.04680}, 2018.

\bibitem{zhuang2020training}
B.~Zhuang, L.~Liu, M.~Tan, C.~Shen, and I.~Reid, ``Training quantized neural networks with a full-precision auxiliary module,'' in \emph{Proceedings of the IEEE/CVF conference on computer vision and pattern recognition}, 2020, pp. 1488--1497.

\bibitem{peng2022training}
F.~Peng, S.~Liu, N.~Lu, and K.~Tang, ``Training quantized deep neural networks via cooperative coevolution,'' in \emph{International Conference on Sensing and Imaging}.\hskip 1em plus 0.5em minus 0.4em\relax Springer, 2022, pp. 81--93.

\bibitem{vaswani2017attention}
A.~Vaswani, N.~Shazeer, N.~Parmar, J.~Uszkoreit, L.~Jones, A.~N. Gomez, {\L}.~Kaiser, and I.~Polosukhin, ``Attention is all you need,'' \emph{Advances in neural information processing systems}, vol.~30, 2017.

\bibitem{hooper2024kvquant}
C.~Hooper, S.~Kim, H.~Mohammadzadeh, M.~W. Mahoney, Y.~S. Shao, K.~Keutzer, and A.~Gholami, ``Kvquant: Towards 10 million context length llm inference with kv cache quantization,'' \emph{arXiv preprint arXiv:2401.18079}, 2024.

\bibitem{aminabadi2022deepspeed}
R.~Y. Aminabadi, S.~Rajbhandari, A.~A. Awan, C.~Li, D.~Li, E.~Zheng, O.~Ruwase, S.~Smith, M.~Zhang, J.~Rasley \emph{et~al.}, ``Deepspeed-inference: enabling efficient inference of transformer models at unprecedented scale,'' in \emph{SC22: International Conference for High Performance Computing, Networking, Storage and Analysis}.\hskip 1em plus 0.5em minus 0.4em\relax IEEE, 2022, pp. 1--15.

\bibitem{sheng2023flexgen}
Y.~Sheng, L.~Zheng, B.~Yuan, Z.~Li, M.~Ryabinin, B.~Chen, P.~Liang, C.~R{\'e}, I.~Stoica, and C.~Zhang, ``Flexgen: High-throughput generative inference of large language models with a single gpu,'' in \emph{International Conference on Machine Learning}.\hskip 1em plus 0.5em minus 0.4em\relax PMLR, 2023, pp. 31\,094--31\,116.

\bibitem{liu2024scissorhands}
Z.~Liu, A.~Desai, F.~Liao, W.~Wang, V.~Xie, Z.~Xu, A.~Kyrillidis, and A.~Shrivastava, ``Scissorhands: Exploiting the persistence of importance hypothesis for llm kv cache compression at test time,'' \emph{Advances in Neural Information Processing Systems}, vol.~36, 2024.

\bibitem{zhang2024h2o}
Z.~Zhang, Y.~Sheng, T.~Zhou, T.~Chen, L.~Zheng, R.~Cai, Z.~Song, Y.~Tian, C.~R{\'e}, C.~Barrett \emph{et~al.}, ``H2o: Heavy-hitter oracle for efficient generative inference of large language models,'' \emph{Advances in Neural Information Processing Systems}, vol.~36, 2024.

\bibitem{liu2024kivi}
Z.~Liu, J.~Yuan, H.~Jin, S.~Zhong, Z.~Xu, V.~Braverman, B.~Chen, and X.~Hu, ``Kivi: A tuning-free asymmetric 2bit quantization for kv cache,'' \emph{arXiv preprint arXiv:2402.02750}, 2024.

\bibitem{dong2024qaq}
S.~Dong, W.~Cheng, J.~Qin, and W.~Wang, ``Qaq: Quality adaptive quantization for llm kv cache,'' \emph{arXiv preprint arXiv:2403.04643}, 2024.

\bibitem{duanmu2024skvq}
H.~Duanmu, Z.~Yuan, X.~Li, J.~Duan, X.~Zhang, and D.~Lin, ``Skvq: Sliding-window key and value cache quantization for large language models,'' \emph{arXiv preprint arXiv:2405.06219}, 2024.

\bibitem{yue2024wkvquant}
Y.~Yue, Z.~Yuan, H.~Duanmu, S.~Zhou, J.~Wu, and L.~Nie, ``Wkvquant: Quantizing weight and key/value cache for large language models gains more,'' \emph{arXiv preprint arXiv:2402.12065}, 2024.

\bibitem{devoto2024simple}
A.~Devoto, Y.~Zhao, S.~Scardapane, and P.~Minervini, ``A simple and effective $ l\_2 $ norm-based strategy for kv cache compression,'' \emph{arXiv preprint arXiv:2406.11430}, 2024.

\bibitem{zuhri2024mlkv}
Z.~M.~K. Zuhri, M.~F. Adilazuarda, A.~Purwarianti, and A.~F. Aji, ``Mlkv: Multi-layer key-value heads for memory efficient transformer decoding,'' \emph{arXiv preprint arXiv:2406.09297}, 2024.

\bibitem{brandon2024reducing}
W.~Brandon, M.~Mishra, A.~Nrusimha, R.~Panda, and J.~R. Kelly, ``Reducing transformer key-value cache size with cross-layer attention,'' \emph{arXiv preprint arXiv:2405.12981}, 2024.

\bibitem{adnan2024keyformer}
M.~Adnan, A.~Arunkumar, G.~Jain, P.~Nair, I.~Soloveychik, and P.~Kamath, ``Keyformer: Kv cache reduction through key tokens selection for efficient generative inference,'' \emph{Proceedings of Machine Learning and Systems}, vol.~6, pp. 114--127, 2024.

\bibitem{kang2024gear}
H.~Kang, Q.~Zhang, S.~Kundu, G.~Jeong, Z.~Liu, T.~Krishna, and T.~Zhao, ``Gear: An efficient kv cache compression recipefor near-lossless generative inference of llm,'' \emph{arXiv preprint arXiv:2403.05527}, 2024.

\bibitem{liu2024minicache}
A.~Liu, J.~Liu, Z.~Pan, Y.~He, G.~Haffari, and B.~Zhuang, ``Minicache: Kv cache compression in depth dimension for large language models,'' \emph{arXiv preprint arXiv:2405.14366}, 2024.

\bibitem{zhang2024cached}
Z.~Zhang, W.~Shao, Y.~Ge, X.~Wang, J.~Gu, and P.~Luo, ``Cached transformers: Improving transformers with differentiable memory cachde,'' in \emph{Proceedings of the AAAI Conference on Artificial Intelligence}, vol.~38, no.~15, 2024, pp. 16\,935--16\,943.

\bibitem{wu2024layer}
H.~Wu and K.~Tu, ``Layer-condensed kv cache for efficient inference of large language models,'' \emph{arXiv preprint arXiv:2405.10637}, 2024.

\bibitem{liu2021post}
Z.~Liu, Y.~Wang, K.~Han, W.~Zhang, S.~Ma, and W.~Gao, ``Post-training quantization for vision transformer,'' \emph{Advances in Neural Information Processing Systems}, vol.~34, pp. 28\,092--28\,103, 2021.

\bibitem{liu2023pd}
J.~Liu, L.~Niu, Z.~Yuan, D.~Yang, X.~Wang, and W.~Liu, ``Pd-quant: Post-training quantization based on prediction difference metric,'' in \emph{Proceedings of the IEEE/CVF Conference on Computer Vision and Pattern Recognition}, 2023, pp. 24\,427--24\,437.

\bibitem{hubara2021accurate}
I.~Hubara, Y.~Nahshan, Y.~Hanani, R.~Banner, and D.~Soudry, ``Accurate post training quantization with small calibration sets,'' in \emph{International Conference on Machine Learning}.\hskip 1em plus 0.5em minus 0.4em\relax PMLR, 2021, pp. 4466--4475.

\bibitem{Banner2018PostT4}
R.~Banner, Y.~Nahshan, and D.~Soudry, ``Post training 4-bit quantization of convolutional networks for rapid-deployment,'' in \emph{Neural Information Processing Systems}, 2018.

\bibitem{Fang2020PosttrainingPL}
J.~Fang, A.~Shafiee, H.~Abdel-Aziz, D.~Thorsley, G.~Georgiadis, and J.~Hassoun, ``Post-training piecewise linear quantization for deep neural networks,'' in \emph{European Conference on Computer Vision}, 2020.

\bibitem{Nagel2020UpOD}
\BIBentryALTinterwordspacing
M.~Nagel, R.~A. Amjad, M.~van Baalen, C.~Louizos, and T.~Blankevoort, ``Up or down? adaptive rounding for post-training quantization,'' \emph{ArXiv}, vol. abs/2004.10568, 2020. [Online]. Available: \url{https://api.semanticscholar.org/CorpusID:216056295}
\BIBentrySTDinterwordspacing

\bibitem{liu2023llm}
Z.~Liu, B.~Oguz, C.~Zhao, E.~Chang, P.~Stock, Y.~Mehdad, Y.~Shi, R.~Krishnamoorthi, and V.~Chandra, ``Llm-qat: Data-free quantization aware training for large language models,'' \emph{arXiv preprint arXiv:2305.17888}, 2023.

\bibitem{shang2023pbllmpartiallybinarizedlarge}
\BIBentryALTinterwordspacing
Y.~Shang, Z.~Yuan, Q.~Wu, and Z.~Dong, ``Pb-llm: Partially binarized large language models,'' 2023. [Online]. Available: \url{https://arxiv.org/abs/2310.00034}
\BIBentrySTDinterwordspacing

\bibitem{shen2024edgeqatentropydistributionguided}
\BIBentryALTinterwordspacing
X.~Shen, Z.~Kong, C.~Yang, Z.~Han, L.~Lu, P.~Dong, C.~Lyu, C.~hsiang Li, X.~Guo, Z.~Shu, W.~Niu, M.~Leeser, P.~Zhao, and Y.~Wang, ``Edgeqat: Entropy and distribution guided quantization-aware training for the acceleration of lightweight llms on the edge,'' 2024. [Online]. Available: \url{https://arxiv.org/abs/2402.10787}
\BIBentrySTDinterwordspacing

\bibitem{ashkboos2024quarot}
S.~Ashkboos, A.~Mohtashami, M.~L. Croci, B.~Li, M.~Jaggi, D.~Alistarh, T.~Hoefler, and J.~Hensman, ``Quarot: Outlier-free 4-bit inference in rotated llms,'' \emph{arXiv preprint arXiv:2404.00456}, 2024.

\bibitem{bondarenko2024lowrankquantizationawaretrainingllms}
\BIBentryALTinterwordspacing
Y.~Bondarenko, R.~D. Chiaro, and M.~Nagel, ``Low-rank quantization-aware training for llms,'' 2024. [Online]. Available: \url{https://arxiv.org/abs/2406.06385}
\BIBentrySTDinterwordspacing

\bibitem{chen2024efficientqatefficientquantizationawaretraining}
\BIBentryALTinterwordspacing
M.~Chen, W.~Shao, P.~Xu, J.~Wang, P.~Gao, K.~Zhang, Y.~Qiao, and P.~Luo, ``Efficientqat: Efficient quantization-aware training for large language models,'' 2024. [Online]. Available: \url{https://arxiv.org/abs/2407.11062}
\BIBentrySTDinterwordspacing

\bibitem{jacob2017quantizationtrainingneuralnetworks}
\BIBentryALTinterwordspacing
B.~Jacob, S.~Kligys, B.~Chen, M.~Zhu, M.~Tang, A.~Howard, H.~Adam, and D.~Kalenichenko, ``Quantization and training of neural networks for efficient integer-arithmetic-only inference,'' 2017. [Online]. Available: \url{https://arxiv.org/abs/1712.05877}
\BIBentrySTDinterwordspacing

\bibitem{esser2020learnedstepsizequantization}
\BIBentryALTinterwordspacing
S.~K. Esser, J.~L. McKinstry, D.~Bablani, R.~Appuswamy, and D.~S. Modha, ``Learned step size quantization,'' 2020. [Online]. Available: \url{https://arxiv.org/abs/1902.08153}
\BIBentrySTDinterwordspacing

\bibitem{gao2022efficientgraphneuralnetwork}
\BIBentryALTinterwordspacing
X.~Gao, W.~Zhang, Y.~Shao, Q.~V.~H. Nguyen, B.~Cui, and H.~Yin, ``Efficient graph neural network inference at large scale,'' 2022. [Online]. Available: \url{https://arxiv.org/abs/2211.00495}
\BIBentrySTDinterwordspacing

\end{thebibliography}
}

\end{document}